\documentclass[]{fairmeta}

\usepackage{cleveref}
\usepackage{textgreek}

\crefname{figure}{Figure}{Figures}
\Crefname{figure}{Figure}{Figures}

\crefname{section}{Section}{Sections}
\Crefname{section}{Section}{Sections}
\crefname{table}{Table}{Tables}
\Crefname{table}{Table}{Tables}

\crefformat{equation}{(#2#1#3)}
\crefmultiformat{equation}{(#2#1#3)}{ and~(#2#1#3)}{, (#2#1#3)}{ and~(#2#1#3)}
\crefrangeformat{equation}{(#3#1#4) to~(#5#2#6)}

\usepackage{xspace}
\usepackage{enumitem}
\usepackage{wrapfig}
\usepackage{threeparttable}

\usepackage{tikz}
\usetikzlibrary{arrows.meta, positioning, fit, calc, backgrounds, shapes.geometric, decorations.pathreplacing}

\usepackage{fancyvrb}

\usepackage{longtable}
\usepackage{array}
\usepackage{colortbl}
\usepackage{pifont}
\newcommand{\cmark}{\textcolor{dashGreenDark}{\ding{51}}}
\newcommand{\xmark}{\textcolor{dashGrey}{\ding{55}}}
\tcbuselibrary{raster}   %
\usepackage{tabularx}

\newcommand{\highlight}[1]{{\color{roleAgent}\textbf{#1}}}

\newtcolorbox{myframe}{colback=metabg, colframe=metabg, boxrule=0pt, arc=5pt,
    left=8pt, right=8pt, top=4pt, bottom=4pt, before skip=7pt, after skip=7pt}
\numberwithin{equation}{section}   %
\renewcommand{\eqref}[1]{\labelcref{#1}}

\definecolor{dashBlue}{HTML}{1877F2}      %
\definecolor{dashOrange}{HTML}{F0701A}    %
\definecolor{dashPurple}{HTML}{5A24C7}    %
\definecolor{dashPink}{HTML}{E42C97}      %
\definecolor{dashCyan}{HTML}{00487C}      %
\definecolor{dashTeal}{HTML}{0EAC96}      %
\definecolor{dashGreen}{HTML}{32AB4F}     %
\definecolor{dashGreenDark}{HTML}{2A9142} %
\definecolor{dashYellow}{HTML}{F7B928}    %
\definecolor{dashRed}{HTML}{E41E3F}       %
\definecolor{dashGrey}{HTML}{8A8D91}      %
\definecolor{dashGreyDark}{HTML}{3A3B3C}  %

\colorlet{roleAgent}{dashBlue}
\colorlet{roleLenz}{dashOrange}
\colorlet{roleExp}{dashGreyDark}   %
\colorlet{roleCtx}{dashGrey}

\colorlet{saraBlue}{dashBlue!16}          %
\colorlet{saraBlueSoft}{dashBlue!7}       %
\colorlet{saraOrange}{dashOrange!16}      %
\colorlet{saraOrangeSoft}{dashOrange!7}   %
\colorlet{saraGrey}{dashGrey!38}          %
\colorlet{saraGreySoft}{dashGrey!15}      %
\colorlet{botorchorange}{roleLenz}        %

\tcbset{dlgbase/.style={
    enhanced, arc=3pt, boxrule=0pt,
    fonttitle=\bfseries\footnotesize\sffamily, fontupper=\footnotesize,
    left=6pt, right=6pt, top=2.5pt, bottom=2.5pt,
    before skip=1.6pt, after skip=1.6pt,
    attach title to upper={\hspace{0.45em}}}}
\newtcolorbox{dlghuman}[1][Human]{dlgbase, colback=saraGreySoft, colframe=saraGrey,
    boxrule=0.4pt, coltitle=black!62, title={#1}}
\newtcolorbox{dlgsara}[1][Sara]{dlgbase, colback=saraBlue, colframe=saraBlue,
    coltitle=roleAgent, title={#1}, right skip=0.32\linewidth}
\newtcolorbox{dlglenz}[1][lenz]{dlgbase, colback=saraOrange, colframe=saraOrange,
    coltitle=botorchorange, title={#1}, fontupper=\ttfamily\scriptsize,
    left skip=0.32\linewidth}

\usepackage{listings}
\definecolor{lzStr}{HTML}{1F7A4D}   %
\definecolor{lzNum}{HTML}{B45309}   %
\definecolor{lzPun}{HTML}{9AA3B2}   %
\definecolor{lzKw}{HTML}{7C3AED}    %
\definecolor{lzframe}{HTML}{D9DEE7} %
\definecolor{lzbar}{HTML}{EEF1F6}   %
\definecolor{lzout}{HTML}{EEF3FC}   %

\lstdefinestyle{lenzconsole}{
  language=lenzcli,
  basicstyle=\ttfamily\scriptsize\color{metafg},
  aboveskip=1pt, belowskip=1pt,
  showstringspaces=false, keepspaces=true, columns=fullflexible,
  breaklines=true, breakatwhitespace=false,
  keywordstyle=[1]\color{roleLenz}\bfseries,   %
  keywordstyle=[2]\color{lzsub}\bfseries,       %
  keywordstyle=[3]\color{lzflag},               %
  stringstyle=\color{lzStr},                    %
  commentstyle=\color{lzcom}\itshape}           %

\lstdefinestyle{lenzjson}{
  basicstyle=\ttfamily\scriptsize\color{metafg},
  aboveskip=1pt, belowskip=1pt,
  escapeinside={(*}{*)},
  showstringspaces=false, keepspaces=true, columns=fullflexible, breaklines=false,
  stringstyle=\color{lzStr}, morestring=[b]",
  morekeywords={true,false,null}, keywordstyle=\color{lzKw},
  literate=
    {0}{{\textcolor{lzNum}{0}}}1 {1}{{\textcolor{lzNum}{1}}}1
    {2}{{\textcolor{lzNum}{2}}}1 {3}{{\textcolor{lzNum}{3}}}1
    {4}{{\textcolor{lzNum}{4}}}1 {5}{{\textcolor{lzNum}{5}}}1
    {6}{{\textcolor{lzNum}{6}}}1 {7}{{\textcolor{lzNum}{7}}}1
    {8}{{\textcolor{lzNum}{8}}}1 {9}{{\textcolor{lzNum}{9}}}1
    {\{}{{\textcolor{lzPun}{\{}}}1 {\}}{{\textcolor{lzPun}{\}}}}1
    {[}{{\textcolor{lzPun}{[}}}1 {]}{{\textcolor{lzPun}{]}}}1
    {:}{{\textcolor{lzPun}{:}}}1 {,}{{\textcolor{lzPun}{,}}}1}

\newtcolorbox{lenzex}{
    enhanced, breakable=false, arc=2.5pt, boxrule=0pt, colframe=metabg,
    colback=metabg,
    left=8pt, right=8pt, top=2.5pt, bottom=3pt, middle=4pt,
    before skip=0pt, after skip=0pt,
    segmentation style={solid, black!22, line width=0.5pt}}

\tcbset{cdlg/.style={enhanced, arc=3.5pt, boxrule=0pt, frame hidden,
    fonttitle=\bfseries\scriptsize\sffamily, fontupper=\scriptsize,
    left=6pt, right=6pt, top=2.5pt, bottom=3pt,
    before skip=0pt, after skip=3.5pt,
    attach title to upper={\hspace{0.5em}}}}
\newtcolorbox{chuman}{cdlg, colback=saraGreySoft, coltitle=black!55, title=Human}
\newtcolorbox{csara}{cdlg, colback=saraBlue, coltitle=roleAgent, title=Sara,
    right skip=0.12\linewidth}
\newtcolorbox{clenz}{cdlg, colback=saraOrangeSoft,
    coltitle=botorchorange!85!black, title=lenz, fontupper=\ttfamily\scriptsize,
    left skip=0.14\linewidth}
\colorlet{saraExpFill}{dashGreyDark!8}
\newtcolorbox{cexp}{cdlg, colback=saraExpFill,
    coltitle=roleExp, title=experiment, fontupper=\ttfamily\scriptsize,
    left skip=0.14\linewidth}
\newtcolorbox{cthink}{enhanced, colback=white, colframe=white, arc=1pt, boxrule=0pt,
    borderline west={1.6pt}{0pt}{roleAgent!70},
    left=1pt, right=1pt, top=1pt, bottom=1pt, before skip=2pt, after skip=3pt,
    fontupper=\scriptsize, height from=5mm to 40mm,
    overlay={\node[rotate=90, anchor=center, font=\bfseries\tiny\sffamily,
        text=roleAgent!75] at ([xshift=-4.5pt]frame.west) {reasoning};}}

\tcbuselibrary{listings} 
\usepackage{listings}

\definecolor{lzsub}{HTML}{2A2B2C}    %
\definecolor{lzflag}{HTML}{2C6E8F}   %
\definecolor{lzcom}{HTML}{8A8D91}    %

\lstdefinelanguage{lenzcli}{
  sensitive=true,
  alsoletter={-},
  morekeywords=[1]{lenz},
  morekeywords=[2]{create,suggest,submit,observe,%
    set-bounds,set-acqf,set-objectives,set-constraints,%
    diagnostics,predict,score,incumbent,pareto,trials},
  morekeywords=[3]{--space,--objectives,--constraints,--acqf,--q,--around,%
    --config,--configs,--metrics,--bounds,--beta,--state},
  morecomment=[l]{\#},
  morestring=[b]',
}

\lstdefinelanguage{pi}{
  sensitive=true,
  alsoletter={-},
  morekeywords=[1]{pi},
  morekeywords=[2]{create,suggest,submit,observe,%
    set-bounds,set-acqf,set-objectives,set-constraints,%
    diagnostics,predict,score,incumbent,pareto,trials},
  morekeywords=[3]{--tools,--system-prompt,--append-system-prompt},
  morecomment=[l]{\#},
  morestring=[b]",
}

\lstdefinestyle{lenz}{
  language=lenzcli, 
  basicstyle=\ttfamily\small\color{metafg},
  backgroundcolor=\color{metabg},
  frame=single, framerule=0.0pt, framesep=6pt,
  xleftmargin=1em, xrightmargin=0.5em,
  breaklines=true, breakatwhitespace=false,
  columns=fullflexible, keepspaces=true, showstringspaces=false,
  keywordstyle=[1]\color{roleLenz}\bfseries,   %
  keywordstyle=[2]\color{lzsub}\bfseries,       %
  keywordstyle=[3]\color{lzflag},               %
  stringstyle=\color{lzStr},                    %
  commentstyle=\color{lzcom}\itshape,           %
}

\lstdefinestyle{sara}{
  language=pi, 
  basicstyle=\ttfamily\small\color{metafg},
  backgroundcolor=\color{metabg},
  frame=single, framerule=0.0pt, framesep=6pt,
  xleftmargin=1em, xrightmargin=0.5em,
  breaklines=true, breakatwhitespace=false,
  columns=fullflexible, keepspaces=true, showstringspaces=false,
  keywordstyle=[1]\color{roleAgent}\bfseries,   %
  keywordstyle=[2]\color{lzsub}\bfseries,       %
  keywordstyle=[3]\color{lzflag},               %
  stringstyle=\color{lzStr},                    %
  commentstyle=\color{lzcom}\itshape,           %
}

\usepackage{listingsutf8}  %

\usepackage{amsmath,amsfonts, amssymb, bm}
\usepackage{mathtools}
\usepackage{amsthm}
\usepackage{thmtools}

\def\1{\bm{1}}

\def\vh{{\bm{h}}}

\def\vx{{\bm{x}}}
\def\vy{{\bm{y}}}

\DeclareMathAlphabet{\mathsfit}{\encodingdefault}{\sfdefault}{m}{sl}
\SetMathAlphabet{\mathsfit}{bold}{\encodingdefault}{\sfdefault}{bx}{n}

\def\gA{{\mathcal{A}}}
\def\gB{{\mathcal{B}}}
\def\gC{{\mathcal{C}}}
\def\gD{{\mathcal{D}}}
\def\gE{{\mathcal{E}}}

\def\gH{{\mathcal{H}}}

\def\gK{{\mathcal{K}}}

\def\gM{{\mathcal{M}}}

\def\gO{{\mathcal{O}}}

\def\gS{{\mathcal{S}}}

\def\gX{{\mathcal{X}}}

\newcommand{\E}{\mathbb{E}}

\newcommand{\R}{\mathbb{R}}

\DeclareMathOperator*{\argmax}{arg\,max}

\DeclareRobustCommand{\sara}{Sara\xspace}
\DeclareRobustCommand{\lenz}{lenz\xspace}

\newcommand{\acqf}{\alpha}               %
\newcommand{\agent}{\mathsf{A}}          %
\newcommand{\policy}{\Pi}                %
\newcommand{\config}{c}                  %
\newcommand{\state}{s}                   %

\newcommand{\cmd}[1]{\texttt{#1}}
\newcommand{\suggest}{\cmd{suggest}}     %
\newcommand{\Aeval}{\mathcal{A}_{\text{eval}}}
\newcommand{\Acomp}{\mathcal{A}_{\text{comp}}}

\usepackage{xspace}

\makeatletter
\DeclareRobustCommand\onedot{\futurelet\@let@token\@onedot}
\def\@onedot{\ifx\@let@token.\else.\null\fi\xspace}

\def\eg{\emph{e.g}\onedot} 
\def\ie{\emph{i.e}\onedot} 
\def\cf{\emph{cf}\onedot}

\makeatother

\title{Agentic Bayesian Optimization through Surrogate-Augmented Autoresearch}
\hypersetup{
  pdftitle={Agentic Bayesian Optimization through Surrogate-Augmented Autoresearch},
  pdfauthor={Paul Brunzema, Louis Tiao, Nhat Le, Kevin De Angeli, Yao Xuan, and Djordje Gligorijevic}
}

\author[1,2,*]{Paul Brunzema}
\author[1]{Louis Tiao}
\author[1]{Nhat Le}
\author[1]{Kevin De Angeli}
\author[1]{Yao Xuan}
\author[1]{Djordje Gligorijevic}

\affiliation[1]{Meta}
\affiliation[2]{RWTH Aachen University, Germany}

\contribution[*]{Work conducted at Meta during an internship}

\abstract{
Bayesian optimization (BO) has become the standard tool for sample-efficient optimization and owes its efficiency to uncertainty-aware search driven by generic statistical priors. 
Richer domain priors can improve BO in principle, but encoding them through tailored kernels or problem structure is difficult and rarely done in practice.
LLMs can help sidestep this difficulty by making informal priors from natural language, code, and documentation directly available to the optimizer.
However, existing LLM-based BO methods either insert the LLM into a fixed role---surrogate, acquisition proxy, or configuration interface---or hand it broad control, sacrificing the systematic exploration that makes BO reliable.
We introduce \highlight{agentic Bayesian optimization}: a paradigm in which an LLM agent is the central decision maker in the BO loop while a Bayesian backend provides the uncertainty-aware optimization substrate.
The agent configures the problem, queries the backend, selects and commits evaluations, and can revise the optimization strategy during the run by tightening bounds, switching acquisition functions, proposing targeted evaluations, or even reframing the problem following new instructions or observed evidence.
We instantiate this idea in \sara, a surrogate-augmented autoresearch agent, and \lenz, a modular BoTorch-based backend that the agent can inspect and modify through a structured interface.
Across synthetic and real-world benchmarks, \sara preserves the reliability of state-of-the-art BO without prior knowledge, outperforms LLM-based baselines, and uses natural-language priors to improve beyond standard BO.
We further demonstrate the practical value of agentic BO in dynamic settings, where \sara reconfigures the full optimization problem on the fly as requirements change---a capability not previously available in standard BO.
}

\date{July 31, 2026}
\correspondence{\email{brunzema@dsme.rwth-aachen.de}, \email{tiao@meta.com}}

\begin{document}

\maketitle

\section{Introduction}
\label{sec:intro}

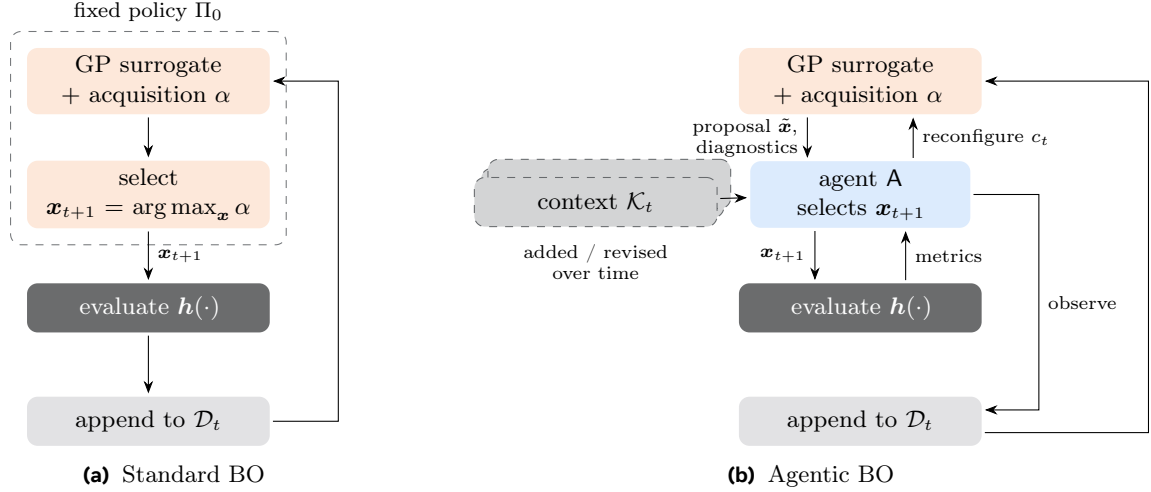
\begin{figure}[t]
    \centering
    \tikzset{
    paradigm box/.style={rounded corners, align=center, minimum height=6.5mm,
                         text width=30mm, inner sep=3pt, fill=dashGrey!25},
    paradigm backend/.style={paradigm box, fill=saraOrange},
    paradigm exp/.style={paradigm box, fill=roleExp!75, text=white},
    paradigm agent/.style={rounded corners, align=center, minimum height=6.5mm,
                           text width=27mm, inner sep=3pt, fill=roleAgent!15},
    paradigm ctx/.style={draw=dashGrey!75!black, dashed, rounded corners, align=center,
                         minimum height=6.5mm, text width=30mm, inner sep=3pt, fill=saraGrey},
    paradigm lbl/.style={font=\footnotesize},
    paradigm elbl/.style={font=\scriptsize},
    paradigm loop/.style={->, shorten >=1pt, shorten <=1pt},
}

\begin{subfigure}[b]{0.35\textwidth}
    \centering
    \begin{tikzpicture}[>=Stealth, font=\small]
        \node[paradigm backend] (apol) at (0, 1.5) {GP surrogate $+$ acquisition $\acqf$};
        \node[paradigm backend] (asel) at (0, 0.0) {select $\vx_{t+1} = \argmax_{\vx} \acqf$};
        \node[paradigm exp] (aeval) at (0,-1.5) {evaluate $\vh(\cdot)$};
        \node[paradigm box] (adata) at (0,-3.0) {append to $\gD_t$};
        \draw[paradigm loop] (apol) -- (asel);
        \draw[paradigm loop] (asel) -- (aeval)
            node[midway, right, paradigm elbl] {$\vx_{t+1}$};
        \draw[paradigm loop] (aeval) -- (adata);
        \draw[paradigm loop] (adata.east) -- ++(0.9,0) |- (apol.east);
        \begin{scope}[on background layer]
            \node[draw=dashGrey!85!black, dashed, rounded corners,
                  fit=(apol)(asel), inner sep=6pt, shorten >=1pt, shorten <=1pt] (apbox) {};
        \end{scope}
        \node[font=\footnotesize] at ($(apbox.north)+(0,0.26)$) {fixed policy $\policy_0$};
    \end{tikzpicture}
    \subcaption{Standard BO}
    \label{fig:paradigm-standard}
\end{subfigure}
\hfill
\begin{subfigure}[b]{0.62\textwidth}
    \centering
    \begin{tikzpicture}[>=Stealth, font=\small]
        \node[paradigm backend] (bpol) at (0, 1.5) {GP surrogate $+$ acquisition $\acqf$};
        \node[paradigm agent] (bagt) at (0, 0.0) {agent $\agent$ \\ selects $\vx_{t+1}$};
        \node[paradigm exp] (beval) at (0,-1.5) {evaluate $\vh(\cdot)$};
        \node[paradigm box] (bdata) at (0,-3.0) {append to $\gD_t$};
        \node[paradigm ctx] (bctx) at (-3.5, -0.1) {context $\gK_t$};
        \begin{scope}[on background layer]
            \node[paradigm ctx] at ([shift={(0.09,0.12)}]bctx.center) {};
            \node[paradigm ctx] at ([shift={(0.18,0.24)}]bctx.center) {};
        \end{scope}
        \node[paradigm elbl, below=1mm of bctx, align=center] {added / revised\\ over time};
        \draw[paradigm loop] ([xshift=-7mm]bpol.south) -- ([xshift=-7mm]bagt.north)
            node[midway, left, paradigm elbl, align=right] {proposal $\tilde{\vx}$,\\ diagnostics};
        \draw[paradigm loop] ([xshift=7mm]bagt.north) -- ([xshift=7mm]bpol.south)
            node[midway, right, paradigm elbl] {reconfigure $\config_t$};
        \draw[paradigm loop] ([xshift=-6mm]bagt.south) -- ([xshift=-6mm]beval.north)
            node[midway, left, paradigm elbl] {$\vx_{t+1}$};
        \draw[paradigm loop] ([xshift=6mm]beval.north) -- ([xshift=6mm]bagt.south)
            node[midway, right, paradigm elbl] {metrics};
        \draw[paradigm loop] (bctx) -- (bagt);
        \draw[paradigm loop] (bagt.east) -- ++(0.9,0) |- ([yshift=1.5mm]bdata.east)
            node[pos=0.25, right, paradigm elbl] {observe};
        \draw[paradigm loop] ([yshift=-1.5mm]bdata.east) -- ++(2.2,0) |- (bpol.east);
    \end{tikzpicture}
    \subcaption{Agentic BO}
    \label{fig:paradigm-agentic}
\end{subfigure}
    \caption{\textit{Standard vs.\ agentic Bayesian optimization.}
    In \cref{fig:paradigm-standard}, a fixed policy with configuration $\config_0$ selects the next query point.
    In \cref{fig:paradigm-agentic}, the agent sits at the center of the loop and decides on the next point---adopting, refining, or overriding the surrogate's proposal $\tilde{\vx}$---while also querying surrogate
    diagnostics and incorporating additional natural-language context or instructions that accrue over the run by reconfiguring the BO backend.}
    \label{fig:paradigm}
\end{figure}

Bayesian optimization (BO) is the standard tool for sample-efficient, zero-order optimization of expensive black-box functions, with applications ranging from hyperparameter tuning and robotics to experimental design~\citep{snoek2012practical,shahriari2016taking,garnett2023bayesian} (\cref{fig:paradigm-standard}).
Its success comes largely from uncertainty-aware sequential decision making: a probabilistic surrogate, typically a Gaussian process~\citep{rasmussen2006gaussian}, tracks what is known about the objective, and an acquisition function uses the posterior to select the next point, trading off exploration and exploitation~\citep{jones1998efficient,srinivas2010gaussian}.
There is, however, a fundamental difference between how standard BO and a human approach the same optimization problem.
In standard BO the configuration---the surrogate, the acquisition function, the search bounds, and the partition of outcomes into objectives and constraints---is fully specified at the start and held fixed throughout, a setting we refer to as a \emph{fixed} policy, whereas a practitioner performing manual optimization adapts the entire search strategy as new evidence arrives.
Approaches that build adaptivity into BO, such as TuRBO~\citep{eriksson2019scalable} or $\pi$BO~\citep{hvarfner2022pibo}, soften this rigidity but do not remove it: their adaptation rules are themselves hand-designed and fixed before the run, so the policy as a whole still does not change in response to what is learned.
Recent interactive BO methods further allow structured user priors to be supplied during a campaign~\citep{seng2025hyperparameter,fehring2025dynamic}.
They enable run-time steering, but fix both the form of feedback and the mechanism by which it affects candidate selection.

A second limitation is that classical BO cannot readily absorb the prior knowledge that domain experts already possess.
Problem-specific kernels can accelerate the search~\citep{marco2017design}, but they are difficult to derive and can capture only a fraction of the rich, heterogeneous knowledge, often expressed in natural language, code, documentation, and past experimental logs, that practitioners actually hold.
Large language models (LLMs) change this picture, as they provide, for the first time, a single model class that can consume all of these sources natively and reason over them jointly, while also deliberating about the optimization as it unfolds.
In principle, this makes LLMs ideally suited to drive adaptive, domain-informed black-box optimization.\footnote{Throughout the paper, we use \emph{black-box optimization} and \emph{zero-order optimization} interchangeably to denote optimization without access to analytical gradients, not to imply a complete absence of problem knowledge.}

In practice, however, current LLM-based methods realize only part of this promise, and they tend to do so in one of two opposing ways.
At one extreme, the LLM replaces the optimization loop entirely.
LLM-as-optimizer methods propose points directly from a textual summary of the history~\citep{yang2024large,zhang2023using,liu2024large,liu2024agenthpo,madiraju2025optimindtune}, and autoresearch agents assume broad control of scientific workflows~\citep{lu2024aiscientist,bran2023chemcrow,boiko2023autonomous,huang2023mlagentbench,zhang2023automlgpt}, thereby gaining adaptivity and access to priors but discarding the explicit probabilistic surrogate and associated uncertainty estimates that have traditionally underpinned BO's sample efficiency in well-defined spaces.
At the other extreme, the LLM is embedded as a single component of an otherwise standard loop, \eg, designing the kernel~\citep{suwandi2026adaptive}, learning surrogate features or fusing predictions into the surrogate~\citep{rankovic2025large,chenlabo}, selecting the acquisition function~\citep{ngo2026adaptive}, proposing or scoring candidate points~\citep{cisse2025language,yang2026reasoningbo}, or translating natural-language feedback and arbitrating between the LLM and the surrogate~\citep{kobalczyk2026lilo,chang2025llinbo}.
All of these preserve a probabilistic surrogate in some form but leave the optimization policy, including all developed heuristics, fixed during optimization.
An agent that can read code, form hypotheses, and reason about the search landscape, by contrast, should be able to change \emph{how} it optimizes---tightening bounds, switching acquisition functions, or reframing the problem mid-run---without surrendering the uncertainty quantification that makes the search efficient.
None of these approaches fully harnesses the best of both worlds: an LLM that excels at sequential decision making together with a surrogate that supplies uncertainty estimates.

We introduce \highlight{agentic Bayesian optimization}, a paradigm in which an LLM agent sits at the center of the optimization loop while delegating probabilistic modeling and acquisition optimization to a Bayesian backend (\cref{fig:paradigm-agentic}).
This division plays to each component's strength: the agent contributes semantic priors and judgments that are difficult to encode in standard BO, while the backend provides posterior uncertainty for systematic, sample-efficient search.
Before committing each expensive evaluation, the agent may inspect the surrogate, request, refine, or override proposals, and reconfigure the optimization problem in response to new evidence or changing instructions.

This design follows a broader lesson from the LLM literature, where models become more effective when they can delegate precise computation to specialized tools~\citep{schick2023toolformer,gao2023pal,chen2023pot}.
The resulting architecture admits two complementary readings.
From the BO perspective, it is \emph{agentic BO}: the agent controls the optimization process while retaining a probabilistic backend.
From the autoresearch perspective, it is \emph{surrogate-augmented autoresearch}: the research agent delegates uncertainty-aware numerical search to an explicit surrogate and optimization engine.
We instantiate both perspectives in \sara, a \underline{S}urrogate-\underline{a}ugmented auto\underline{r}esearch \underline{a}gent, and \lenz (pronounced [l\textepsilon ns], as in ``lens''), a modular BoTorch-based backend designed for run-time adaptation.

\paragraph{Contributions.}
In summary, the core contributions of this paper are
\begin{enumerate}[label=\textit{(\roman*)}]
  \item \textbf{The paradigm of agentic Bayesian optimization.} We introduce \emph{agentic BO}, in which an agent deliberates over trial data, surrogate diagnostics, and external context to revise the optimization policy during a campaign.
  \item \textbf{A surrogate-augmented autoresearch system.}
  We present \sara and \lenz, which together instantiate agentic BO.
  \sara is an LLM agent that drives the optimization process, while \lenz is a modular BO backend that exposes BO primitives through a structured command-line interface.
  \item \textbf{An empirical study of agentic BO.} Across synthetic and real-world benchmarks, \sara matches state-of-the-art BO performance without prior knowledge, outperforms LLM-based BO methods such as LLAMBO, and uses natural-language priors to improve beyond Ax.
  We further demonstrate its practical value in dynamic settings, where \sara reconfigures the optimization problem as requirements change.
\end{enumerate}

\section{Background and Problem Setting}
\label{sec:background}

We first introduce the class of problems we consider and then frame standard Bayesian
optimization as a fixed policy defined before the campaign begins.

\paragraph{Problem.}
We aim to optimize an expensive-to-evaluate black-box function over a domain $\gX \subseteq \R^d$.
Each evaluation yields a vector of noisy outcomes $\vy_i = \vh(\vx_i) + \bm{\varepsilon}_i$, where $\vh : \gX \to \R^p$, and after $t$ evaluations the collected data is $\gD_t = \{(\vx_i, \vy_i)\}_{i=1}^{t}$.
A problem specification assigns components of $\vh$ to objective functions $f_o$, indexed by $o \in \gO$, and constraint functions $c \in \gC$, with each constraint expressed as $c(\vx) \le 0$.
This gives the constrained multi-objective problem
\begin{equation}
    \label{eq:problem}
    \max_{\vx \in \gX}\; \bigl(f_o(\vx)\bigr)_{o \in \gO}
    \qquad \text{s.t.}\qquad c(\vx) \le 0 \;\;\; \forall\, c \in \gC,
\end{equation}
with the maximum understood in the Pareto sense. 
The familiar cases are recovered by choice of $\gO$ and $\gC$, \ie single-objective BO as $|\gO|=1$, $\gC = \emptyset$, constrained BO as $\gC \neq \emptyset$, and multi-objective BO as $|\gO| > 1$. 
Which outcomes serve as objectives and which as constraints is a modeling choice, not a property of the experiment itself.

\paragraph{The standard BO loop.}
Bayesian optimization fits a probabilistic surrogate to $\gD_t$ and uses it to select the next query.
Placing a GP on each outcome gives a posterior with mean $\mu_t$ and variance $\sigma_t^2$~\citep{rasmussen2006gaussian}, and an acquisition function $\acqf(\vx; \gD_t)$ scores candidate points by their expected utility under this posterior.
It is useful to separate two steps that the standard loop conflates: the acquisition function \emph{proposes} by ranking $\gX$, and a selection rule \emph{commits} one candidate as the next evaluation.
Standard BO uses the $\argmax$ rule,
\begin{equation}
    \label{eq:standard-bo}
    \vx_{t+1} = \argmax_{\vx \in \gB}\, \acqf(\vx; \gD_t),
\end{equation}
over an active search region $\gB \subseteq \gX$. 
We keep these two steps distinct in what follows, since an agent can retain the surrogate's proposal while overriding the selection.

\paragraph{Policy and configuration.}
More broadly, the full set of choices active at a given campaign step---the model class
$\gM_t$, acquisition function $\acqf_t$, active region $\gB_t$, and problem partition
$(\gO_t,\gC_t)$---form the \highlight{configuration}
\begin{equation}
  \label{eq:config}
  \config_t = (\gM_t,\,\acqf_t,\,\gB_t,\,\gO_t,\,\gC_t).
\end{equation}
A standard BO loop similar to the one described above then amounts to a fixed policy $\policy_0$ that maps the accumulated data to the active configuration and next query,
\begin{equation}
  \label{eq:standard-bo-policy}
  (\config_t,\vx_{t+1}) = \policy_0(\gD_t),
  \qquad
  \policy_0 \text{ fixed before the campaign}.
\end{equation}
Here, \emph{fixed} refers to the mapping $\policy_0$, but not its outputs, \ie, the configuration may change during the campaign, but only according to rules prescribed in advance.
Modern BO frameworks~\citep{olson2025ax,akiba2019optuna} are concrete instances of such a policy.
For example, Ax switches from Sobol initialization to model-based search, but both the switching rule and the search strategy are determined before the first evaluation; similarly, most specialized BO algorithms are fully specified by heuristics at startup.

\section{Related Work}
\label{sec:related}

Our work sits at the intersection of BO, automated machine learning (AutoML), LLM-driven optimization, and autonomous research agents.
We organize prior work by whether and where the LLM enters the optimization loop: not at all (classical BO and AutoML), as a single \emph{fixed} component of it (hybrid LLM--BO), \emph{in place of} it (LLMs as the optimizer), or \emph{around} a broader workflow (open-ended agents).
\Cref{tab:landscape} organizes these by whether decisions can be supported by a probabilistic surrogate, how much run-time control the method exerts over its own policy, and whether it can consume natural-language priors and call tools.

\begin{table}[t]
    \centering
    \small
    \setlength{\tabcolsep}{6pt}
    \renewcommand{\arraystretch}{1.15}
    \begin{tabular}{@{}l ccc@{}}
        \toprule
        & \shortstack{Calibrated\\ surrogate} & \shortstack{Full run-time\\ policy control} & \shortstack{Natural-language\\ priors \& tool use} \\
        \midrule
        Classical BO \& AutoML {\scriptsize (Ax, TPE, OptFormer, \dots)}    & \cmark & \xmark & \xmark \\
        Interactive BO {\scriptsize (IBO-HPC, DynaBO, \dots)} & \cmark & \xmark & \xmark \\
        Hybrid LLM-BO {\scriptsize (LILO, BORA, Reasoning BO, \dots)}       & \cmark & \xmark & \cmark \\
        LLM-as-optimizer {\scriptsize (LLAMBO, OPRO, AgentHPO, \dots)}             & \xmark & \cmark & \cmark \\
        Open-ended autoresearch agents {\scriptsize (HyperAgents, \dots)}                & \xmark & \cmark & \cmark \\
        \midrule
        \rowcolor{roleAgent!10}
        \textbf{Agentic BO / Surrogate-augmented autoresearch} {\scriptsize (\sara\ + \lenz)}                  & \cmark & \cmark & \cmark \\
        \bottomrule
    \end{tabular}
    \caption{\textit{Agentic BO in the optimization landscape.}
    Classical BO and AutoML ground every decision in a probabilistic surrogate but run a \emph{fixed} policy.
    LLM-as-optimizer methods and open-ended autoresearch agents exercise run-time control but discard posterior uncertainty.
    \sara retains access to a BO backend \lenz \emph{and} reasons about the search.}
    \label{tab:landscape}
\end{table}

\subsection{Bayesian optimization and AutoML}

Classical BO couples a probabilistic surrogate, typically a GP~\citep{rasmussen2006gaussian}, with an acquisition function that balances exploration and exploitation~\citep{jones1998efficient, srinivas2010gaussian, shahriari2016taking, frazier2018tutorial,garnett2023bayesian}.
Mature frameworks such as Ax~\citep{olson2025ax} or Optuna~\citep{akiba2019optuna} package these components into reusable routines, but by design the optimization policy is declared before the campaign and held fixed throughout.
Notable extensions equip BO with adaptive mechanisms: TuRBO~\citep{eriksson2019scalable,eriksson2021scalable} restricts the search to a local region that expands or contracts with progress, while user priors can bias early search through the acquisition function~\citep{hvarfner2022pibo}.
Both are \emph{hand-designed} adaptation rules for a specific situation.
Interactive BO extends this idea by allowing users to steer the search during a campaign.
IBO-HPC conditions candidate generation on user-specified beliefs through probabilistic circuits~\citep{seng2025hyperparameter}, while DynaBO incorporates successive priors and safeguards against misleading ones~\citep{fehring2025dynamic}.
Both retain a probabilistic surrogate but restrict interaction to structured priors interpreted by a fixed selection rule.
Agentic BO broadens this interaction by allowing an agent to reason over unstructured context and the evolving campaign, and to adapt the optimization process accordingly.
Beyond BO, the broader AutoML literature---from random search~\citep{bergstra2012random} and tree-structured estimators~\citep{bergstra2011tpe} to multi-fidelity methods~\citep{li2018hyperband} and meta-learned optimizers~\citep{chen2022optformer}---similarly commits to a fixed search strategy in advance~\citep{tornede2024automl}.
\sara keeps the explicit probabilistic model but inverts the control relationship: \emph{which} strategy to apply, and \emph{when}, becomes a decision made by reasoning over the run, informed by the priors extracted from the instructions and the observations gathered during it.

\subsection{Hybrid LLM--BO methods}

A growing line of work keeps the Bayesian machinery but hands a \emph{single, fixed} component of the loop to an LLM; methods differ in which component that is.
Some target the \emph{kernel}, using an LLM to design or adapt the GP covariance from a contextual description of the problem~\citep{suwandi2026adaptive}.
Others reshape the \emph{surrogate} itself, training LLM embeddings through the GP marginal likelihood so that uncertainty stays calibrated~\citep{rankovic2025large}, or fusing computationally cheap LLM predictions with real evaluations in a multi-fidelity surrogate~\citep{chenlabo}.
A third line acts at the \emph{acquisition} step, letting the LLM select the acquisition function from a portfolio at each iteration~\citep{ngo2026adaptive}.
A fourth group acts on candidate selection, letting the LLM propose or filter candidates~\citep{cisse2025language, yang2026reasoningbo} or arbitrate between LLM and GP proposals~\citep{chang2025llinbo}. 
Finally, \citet{kobalczyk2026lilo} intervene on outcome interpretation, translating natural-language feedback into preferences for a preference GP.
\citet{agarwal2025searching} go further, running BO in a latent embedding space where acquisition proposals bias in-context examples for LLM generation, effectively turning the LLM into a conditioned decoder from surrogate proposals to candidate text.
Together these results show that an LLM and a probabilistic model are complementary, and several report strong gains in the early, low-data regime.
In every case, however, the point of intervention---and the policy around it---is fixed before the first evaluation and the LLM essentially fills one slot in a standard pipeline.
Even the methods that invoke the LLM adaptively do so through a hand-designed trigger, \eg a heuristic on GP uncertainty and performance plateaus~\citep{cisse2025language}, or a fixed probabilistic switching schedule~\citep{mahammadli2024sequential,chang2025llinbo}, rather than leaving the model itself to decide when to intervene.
Agentic BO instead places the agent \emph{at the center} of the loop, where it can reconfigure any component---bounds, objectives, constraints, acquisition behavior, even the objective/constraint partition---at run time, while delegating calibrated numerical search to \lenz.
Importantly, the methods above are complementary to, not superseded by, agentic BO, since an LLM-designed kernel~\citep{suwandi2026adaptive}, a latent-space generation strategy~\citep{agarwal2025searching}, or a preference surrogate~\citep{kobalczyk2026lilo} can each be wrapped as a backend capability that the agent invokes when appropriate, and the system prompt can incorporate the acquisition selection strategy of \citet{ngo2026adaptive}.
The contribution of the agentic layer is orthogonal; it decides \emph{when} and \emph{how} to combine such capabilities in response to accumulating evidence, rather than fixing the choice before the first evaluation.

\subsection{LLMs as the optimizer}

At the other extreme, the optimization itself moves inside the LLM, an approach advocated as a general direction for black-box problems~\citep{song2024position}.
OPRO~\citep{yang2024large} and related methods prompt the model with a trajectory summary and ask it to emit the next configuration directly~\citep{yang2024large, zhang2023using}; program-search methods evolve solutions from LLM-proposed edits~\citep{romeraparedes2024mathematical}; and LLAMBO~\citep{liu2024large} frames this as BO by prompting for a promising point.
None maintains an explicit probabilistic surrogate, so their search is not guided by posterior uncertainty over the objective.
While LLM processes~\citep{requeima2024llmprocesses} show that LLMs can form predictive distributions, a large body of work finds that their uncertainties can be poorly calibrated~\citep{kadavath2022language, tian2023just, xiong2024can}, including for molecular BO specifically~\citep{kristiadi2024sober}.
Recent evidence reinforces the value of a dedicated optimization substrate.
\citet{ferreira2026can} find that classical HPO methods (CMA-ES, TPE) outperform pure-LLM optimizers on a language-model tuning task, and similar findings question whether LLMs are ready for scientific BO on their own~\citep{gupta2025llms}.
On this basis, \citet{ferreira2026can} build Centaur, a hybrid that shares CMA-ES internal state with an LLM through its prompt.
Centaur is the closest baseline in spirit to agentic BO; however, augmenting a prompt underutilizes the sequential decision-making of LLM agents.
We instead give \sara full control over a BO backend and show substantially better performance than Centaur in our experiments.

\subsection{LLM agents for optimization and ML research}

A rapidly growing line of work wraps LLMs in agent loops to drive ML workflows.
AgentHPO~\citep{liu2024agenthpo} reads a task description, proposes configurations, and refines them from past trials; OptiMindTune~\citep{madiraju2025optimindtune} splits the job across cooperating agents; others orchestrate whole pipelines~\citep{zhang2023automlgpt}, generate features~\citep{hollmann2023caafe}, or search architectures by evolving code~\citep{chen2023evoprompting}, and benchmarks now probe how well LLMs act as ML research agents~\citep{huang2023mlagentbench}.
Like the in-context methods above, these drive search by prompting over text summaries of past trials with no calibrated posterior.
Agentic BO differs in that a calibrated surrogate can inform every decision the agent makes.
We believe that agentic BO could be readily incorporated in the above approaches.
More broadly, general reasoning-and-acting frameworks~\citep{yao2023react}, tool-use paradigms~\citep{schick2023toolformer}, multi-agent systems~\citep{hong2023metagpt}, open-ended embodied agents~\citep{wang2023voyager}, and end-to-end scientific discovery pipelines~\citep{lu2024aiscientist, boiko2023autonomous, bran2023chemcrow} emphasize broad autonomy.
A recurring lesson is that LLMs are unreliable at direct computation but effective when they \emph{delegate} to precise tools~\citep{gao2023pal, chen2023pot, schick2023toolformer}.
\sara applies this principle to zero-order optimization, informing her search with a BO backend while remaining deliberately narrower than open-ended agents. %

\section{Agentic Bayesian Optimization}
\label{sec:agentic}

To our knowledge, no existing approach combines three capabilities: \textit{(i)}~retaining a probabilistic surrogate, \textit{(ii)}~adapting the optimization strategy in response to accumulating evidence, and \textit{(iii)}~interpreting and incorporating unstructured domain knowledge or changing task instructions during the campaign.

We introduce agentic BO as an implementation-agnostic framework that combines these capabilities by placing an agent in control of a BO backend.
The framework specifies the information available to the agent, the computational and evaluation actions it may take, and how these actions update or reconfigure the optimization campaign.
Formally, agentic BO extends the fixed-policy loop in~\cref{eq:standard-bo-policy} by replacing its prescribed acquisition step with an agent-controlled \emph{deliberation process} over a Bayesian backend.
Before committing an expensive evaluation, the agent may inspect the surrogate, request and refine proposals, or persistently reconfigure the optimization strategy.
We model this interaction as a \emph{metalevel decision process}, inspired by metalevel MDPs~\citep{hay2012selecting,callaway2022rational}.
The formulation captures two coupled decisions: which computational actions to take and which point to evaluate.
Throughout, \emph{deliberation} refers to the observable sequence of computational actions and backend responses. %
In~\cref{app:meta-mdp}, we show how our formulation can be completed into a full meta-MDP that could form the basis for training or fine-tuning a specialized agentic BO policy.
The goal of this paper, however, is to show that a general-purpose LLM can already implement an effective agentic BO policy without specialized post-training.

\paragraph{State.}
Let $t$ index expensive evaluations and $j$ index computational actions taken between evaluations.
During deliberation, the agent observes the state
\begin{equation}
    \label{eq:state}
    \state_t^{(j)} = \bigl(\gD_t,\; \config_t^{(j)},\; \gK_t,\; \gH_t^{(j)}\bigr),
\end{equation}
where $\gD_t$ is the observed trial data, $\config_t^{(j)}$ is the current optimization configuration (\cf~\cref{eq:config}), $\gK_t$ is the exogenous context, and $\gH_t^{(j)}$ is the deliberation history available to the agent.
The context may include the problem description, domain knowledge, and requirements supplied by an instructor, such as a human or another agent.
It is append-only: new information or revised requirements are added during the campaign, so $\gK_{t+1} \supseteq \gK_t$, with later instructions able to semantically supersede earlier ones.
The history records the retained conversation, reasoning trace, tool calls and responses, proposals, and past interventions.
Although separated conceptually, messages comprising~$\gK_t$ also appear in the conversational history~$\gH_t^{(j)}$ in implementations such as~\sara (\cf~\cref{sec:system}).
Other instantiations may instead expose this context through external memory that the agent accesses during deliberation.

\paragraph{Action space.}
At each deliberation step, the agent chooses among three classes of actions,
\begin{equation}
    \label{eq:action-space}
    \gA = \Acomp \cup \Aeval \cup \{\text{stop}\},
\end{equation}
which differ in their effect on the state and the black-box evaluation budget:

\begin{enumerate}[label=\textit{(\roman*)}]
    \item \textbf{Evaluate} ($\Aeval$).
    The agent commits a point $\vx_{t+1} \in \gX$ to the expensive oracle.
    This is the only action that advances the campaign counter~$t$ and consumes black-box evaluation budget.

    \item \textbf{Computational} ($\Acomp$).
    Backend interactions in this class do \emph{not} trigger an expensive evaluation and can inform the agent's next decision.
    These actions are free with respect to the black-box evaluation budget and do not advance~$t$, although they incur computational overhead (tokens, context length, and backend compute).\footnote{We report token usage in~\cref{app:token_usage} but do not optimize for computational cost in this work.
    A full meta-MDP may account for these computational costs explicitly, as discussed in~\cref{app:meta-mdp}.}
    We further partition them into three main classes:
    \begin{itemize}
        \item \highlight{probe}: read-only queries to the backend---predictions, acquisition scores, diagnostics, the current Pareto front, or trial history---whose responses are added to~$\gH_t^{(j)}$;
        \item \highlight{reconfigure}: mutations of~$\config_t^{(j)}$, \eg, tightening the search region~$\gB$, swapping the acquisition function~$\acqf$, or revising objectives~$\gO$ and constraints~$\gC$;
        \item \highlight{propose}: requests for one or more candidate points $\tilde{\vx}$ from the surrogate backend via acquisition-function optimization over the current active region.
    \end{itemize}

    \item \textbf{Stop} ($\text{stop}$).
    Terminate the campaign and return the incumbent feasible solution in the single-objective case or the current feasible Pareto set in the multi-objective case.
\end{enumerate}

A computational action returns an observation $o_t^{(j)}$ from the backend, and its call and response are appended to the deliberation history.
A reconfiguration additionally updates $\config_t^{(j)} \mapsto \config_t^{(j+1)}$.
None of these operations changes~$\gD_t$ or advances the evaluation counter~$t$.
Deliberation is therefore broader than reconfiguration: probe and propose actions can change the eventual evaluation decision without mutating~$\config_t^{(j)}$, whereas a reconfigure action persistently changes how subsequent backend calls are interpreted or optimized.

\begin{wrapfigure}[16]{r}{0.44\textwidth}
    \centering
    \vspace{-2em}
    \begin{tikzpicture}[
    >=Stealth, font=\small, node distance=0pt,
    state/.style={circle, fill=roleAgent!10, thick,
                  inner sep=1pt, minimum size=0.88cm},
    snext/.style={circle, fill=roleAgent!10, line width=1pt,
                  inner sep=1pt, minimum size=0.88cm},
    init/.style={rounded corners=3pt, fill=roleAgent!10,
                 thick, align=center, inner sep=3.5pt},
    term/.style={circle, draw=dashRed, fill=dashRed!10, double, double distance=1.2pt,
                 line width=0.8pt, text=dashRed!85!black, inner sep=1pt,
                 minimum size=0.9cm, font=\small\sffamily\bfseries},
    expr/.style={rounded corners=3pt, fill=roleExp!75, text=white,
                 font=\small, align=center, inner sep=4pt},
    comp/.style={->, thick, roleLenz, shorten >=1pt, shorten <=1pt},
    complbl/.style={text=roleLenz!88!black, inner sep=2pt},
    eval/.style={->, thick, roleExp, shorten >=1pt, shorten <=1pt},
    stopar/.style={->, thick, dashRed, shorten >=1pt, shorten <=1pt},
    ctxar/.style={->, semithick, shorten >=1pt, shorten <=1pt}, %
    loopback/.style={->, semithick, dashed, shorten >=1pt, shorten <=1pt}, %
    ctxbox/.style={rounded corners=3pt, draw=dashGrey!75!black, dashed,
                   fill=saraGreySoft, inner sep=3.5pt, text=dashGrey!88!black},
]
\node[state] (s0) at (0,0)     {$\state_t^{(0)}$};
\node[state] (s1) at (1.633,0)  {$\state_t^{(1)}$};
\node[state] (sk) at (4.9,0)  {$\state_t^{(k)}$};

\begin{scope}[on background layer]
  \node[rounded corners=6pt, fill=saraOrangeSoft, %
        fit=(s0)(sk), inner sep=8pt] (delib) {};
\end{scope}
\node[text=roleLenz!88!black, anchor=south]
      at ([yshift=0pt, xshift=-15pt]delib.north) {\textbf{deliberation} ($a_k \in \Acomp$)};

\node[init] (setup) at (0,1.9)
     {\highlight{setup}\\[-0.5pt]{\footnotesize\itshape\color{black!55}$\gK_0 \Rightarrow \gX,\gO,\gC,\config_0$}};
\draw[->, semithick, shorten >=1pt, shorten <=1pt] ($(setup.north)+(0,0.45)$) -- (setup.north);
\draw[->, semithick=, shorten >=1pt, shorten <=1pt] (setup) -- (s0);

\node (cd) at (3.266,0) {$\cdots$};
\draw[comp] (s0) -- node[complbl, above] {$a_1$} (s1);
\draw[comp] (s1) -- node[complbl, above] {$a_2$} (cd);
\draw[comp] (cd) -- node[complbl, above] {$a_k$} (sk);

\node[term] (stop) at (4.9,2) {stop};
\draw[stopar] (sk) -- node[above, inner sep=1pt, fill=white, yshift=-3pt] {$a_{k+1}=\text{stop}$} (stop);

\node[expr]  (exp) at (4.9,-1.95) {\textbf{experiment} \\ $\vh(\cdot)$};
\node[snext] (sn)  at (0,-1.95) {$\state_{t+1}^{(0)}$};
\draw[eval] (sk) -- node[left, inner sep=2pt, yshift=-3pt] {$a_{k+1}\!\in\!\Aeval$} (exp);
\draw[eval] (exp) -- node[above, inner sep=2pt] {$\vy_{t+1}$} (sn);

\draw[loopback] (sn) to[out=155,in=205] (s0);

\node[ctxbox] (ktxt) at (1.2,-1.2) {$\gK_{t+1}$};
\draw[ctxar] (ktxt) to[out=180,in=60] (sn);

\end{tikzpicture}
    \caption{\textit{One campaign step in agentic BO.}
    Computational actions inform or reconfigure the search before the agent evaluates or stops.}
    \label{fig:meta-mdp}
\end{wrapfigure}
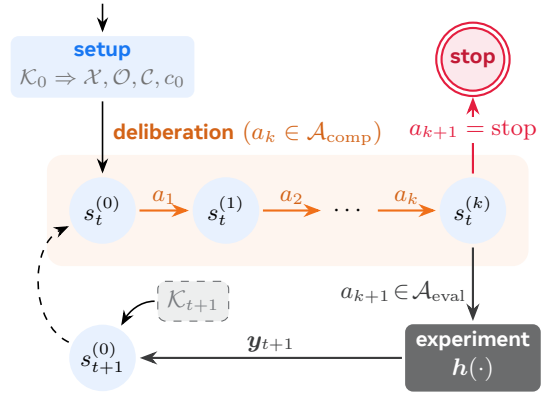

\paragraph{Metalevel policy and deliberation.}
The agent implements a metalevel policy
$
    \agent : \gS \to \gA,
$
which is applied repeatedly to the evolving deliberation state~$\state_t^{(j)}$ until an evaluation or stop action is selected.
Each computational action and its response update the state on which the next action is conditioned, allowing the tool-use strategy to adapt within a campaign step.
A campaign step consists (\cf~\cref{fig:meta-mdp}) of $k_t \ge 0$ computational actions followed by an evaluation or termination:
\begin{equation}
    \label{eq:deliberation}
    a_t^{(0)},\ldots,a_t^{(k_t-1)} \in \Acomp,
    \quad
    a_t^{(k_t)} \in \Aeval \cup \{\text{stop}\}.
\end{equation}
If $a_t^{(k_t)} = \vx_{t+1} \in \Aeval$, the oracle returns $\vy_{t+1} = \vh(\vx_{t+1}) + \bm{\varepsilon}_{t+1}$ and the campaign advances as
\begin{equation}
    \label{eq:state-update}
    \gD_{t+1} = \gD_t \cup \{(\vx_{t+1},\vy_{t+1})\},
    \qquad
    \config_{t+1}^{(0)} = \config_t^{(k_t)}.
\end{equation}
New exogenous information may then be appended to form~$\gK_{t+1}$, and the deliberation history carries forward to the next campaign step.
If instead $a_t^{(k_t)} = \text{stop}$, the campaign terminates.

We additionally distinguish an initial \highlight{setup} phase ($t=0$) in which the agent formalizes the problem from~$\gK_0$, \ie, it defines~$\gX$, partitions outcomes into~$\gO$ and~$\gC$, and initializes~$\config_0^{(0)}$.
We write the complete deliberation induced by~$\agent$ as the campaign-level mapping
\begin{equation}
    \label{eq:agentic}
    (\config_{t+1}^{(0)}, \vx_{t+1}) = \operatorname{Rollout}(\agent,\,\state_t^{(0)})
\end{equation}
with the understanding that this subsumes the deliberation sequence~\cref{eq:deliberation} and may instead return~$\text{stop}$.
Ultimately, it is the \emph{agent} that commits~$\vx_{t+1}$.
To form this decision, it may obtain one or more candidate proposals~$\tilde{\vx}$ from the surrogate and then adopt, refine, or override them.

This formulation includes several familiar regimes as special cases:
\begin{itemize}
    \item without backend interactions, the agent proposes evaluations directly, recovering LLM-as-optimizer methods~\citep{liu2024large,liu2024agenthpo};
    \item with a fixed configuration and one surrogate proposal that is always accepted per campaign step, it recovers the standard BO loop~\cref{eq:standard-bo-policy};
    \item with probes, reconfigurations, multiple proposals, or agent overrides, it yields full agentic BO.
\end{itemize}
The full interaction provides two levels of control beyond the standard loop (also see \Cref{fig:paradigm}).
At the \emph{metalevel}, the agent chooses a response-conditioned sequence of computational actions rather than invoking a prescribed acquisition rule once; this includes, but does not require, making $\config_t^{(j)}$ time-varying through reconfiguration.
At the \emph{evaluation level}, the final decision on $\vx_{t+1}$ rests with the agent rather than a fixed $\argmax$ rule.
Consequently, agentic adaptation can be visible in changing probe and proposal patterns even when the persistent optimization configuration of the backend remains fixed.

The formulation is deliberately agnostic to implementation choices.
It prescribes neither the agent's instantiation, nor the surrogate class, nor how~$\gK_t$ is produced or delivered.
Any system in which the agent observes~$\state_t^{(j)}$, chooses computational actions through~$\Acomp$, and ultimately commits an evaluation query---while retaining or changing the configuration---is an instance of agentic BO.
The next section describes one such instantiation built around a reasoning LLM and a BO backend.

\section{\sara and \lenz as the surrogate-augmented research team}
\label{sec:system}

\begin{figure}[t]
    \centering
    \includegraphics[width=\linewidth]{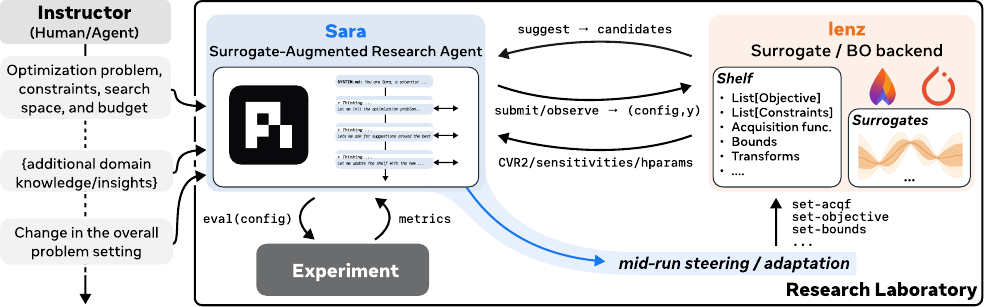}
     \caption{\textit{Our proposed instantiation of agentic Bayesian optimization.} \sara sits at the center of the optimization loop: she reads trial data and surrogate diagnostics, incorporates natural-language context, queries the surrogate, adapts the optimization policy mid-run, and commits the next evaluation.}
    \label{fig:teaser}
\end{figure}

The paradigm of \cref{sec:agentic} requires two ingredients: a flexible backend that the agent can query, steer, and reconfigure without losing data, and an agent that knows when to use each capability.
We instantiate the backend as \lenz (\cref{sec:lenz}), a modular system based on GPyTorch~\citep{gardner2018gpytorch} and BoTorch~\citep{balandat2020botorch}.\footnote{GPyTorch (\url{https://gpytorch.ai/}) and BoTorch (\url{https://botorch.org/}) are available under the MIT license.}
Its interface supports setup and the three computational action classes of agentic BO.
We instantiate the agent as \sara (\cref{sec:sara}), a reasoning LLM that drives the optimization and uses \lenz's surrogates and BO engine, as shown in \cref{fig:teaser}.

\subsection{\lenz: a highly modular backend for agentic Bayesian optimization}
\label{sec:lenz}

\begin{figure}[h]
    \centering
    \includegraphics[width=\linewidth]{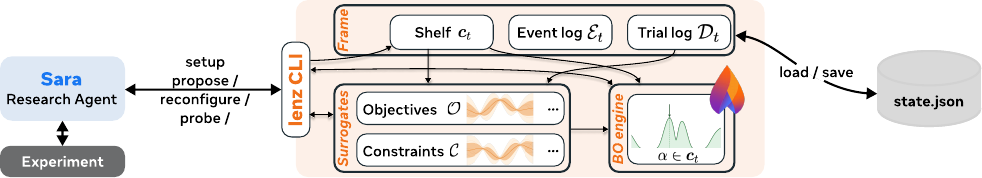}
    \caption{\textit{Schematic overview of \lenz.} All internal functionality is exposed through a single CLI.}
    \label{fig:lenz-frame}
\end{figure}

For \cref{eq:agentic} to be practical, the backend must support the full deliberation phase, \ie probing, proposing, and reconfiguring, without invalidating prior evaluations.
Changing $\config_t$ should reuse every prior evaluation rather than restart the campaign, and the interface must be intuitive enough for an agent to operate from a single reference sheet.
\lenz achieves this by making the raw trial log the single source of truth and exposing a command-line interface (CLI) to the surrogates and to established BO routines. Adding a new constraint metric, tightening bounds, or switching the objective therefore never invalidates collected data.
\Cref{fig:lenz-frame} sketches the resulting architecture.
A \emph{Frame} holds the trial log $\gD_t$, the event log $\gE_t$, and the mutable \emph{shelf} containing the configuration $\config_t$, while the GP surrogates are derived objects rebuilt from $\gD_t$ on each query.
Because the state file remains the sole source of truth, \lenz can keep its engine resident in a per-study daemon with no risk of cache incoherence.
We next describe this CLI in detail.

\paragraph{Interface design.} \lenz exposes its BO primitives through a small, role-structured CLI organized around the action categories introduced in \cref{sec:agentic}.
Below we list the main commands accessible to an agent, grouped by the role they play in the agentic BO loop.
\begin{enumerate}[label=\textit{(\roman*)}]

\item \highlight{setup} (\texttt{create}).
Before the loop begins, the agent declares the search space, objectives, constraints, and initial acquisition function based on the initial instructions $\gK_0$:
\begin{lstlisting}[style=lenz]
lenz create --space '{"lr":{"kind":"range","lower":1e-4,"upper":1e-1,"log_scale":true}, ...}'
               --objectives '{"loss":"minimize"}'
               --constraints '[{"metric":"flops","upper":500}]'
               --acqf noisy_logei
\end{lstlisting}
This materializes the initial configuration $\config_0$.

\item \highlight{propose and evaluate} (\texttt{suggest}, \texttt{submit}, \texttt{observe}).
The agent can request candidates from the surrogate via \texttt{suggest}, which may be called multiple times without consuming evaluation budget.
When the agent commits a configuration for external evaluation, \texttt{submit} marks it as \emph{pending}; subsequent \texttt{suggest} calls fantasize an outcome for that configuration to ensure batch diversity.
After the experiment, \texttt{observe} records the returned metrics in the trial log.
\begin{lstlisting}[style=lenz]
lenz suggest --q N  # suggest a batch of points
lenz suggest --around '{"lr":0.1, "dropout":{"fix":0.0}, ...}'  # around incumbent
lenz suggest --bounds '{"lr":[0.001,0.1],"x2":[10,20]}'  # within provided bounds
lenz submit --config '{...}'  # commit to a run for fantasization
lenz observe --config '{...}' --metrics '{...}'  # record metrics
\end{lstlisting}
Because \texttt{suggest} only queries the surrogates and BO engine, the agent can compare hypothetical proposals under different bounds or acquisition functions before committing.

\item \highlight{reconfigure} (\texttt{set-bounds}, \texttt{set-acqf}, \texttt{set-objectives}, \texttt{set-constraints}).
The agent can edit the full specification of the optimization problem at every time step:
\begin{lstlisting}[style=lenz]
lenz set-bounds --bounds '{"lr":[1e-4,1e-3]}'  # refine search bounds
lenz set-acqf --acqf ucb --beta 2.0  # change acquisition function (and its hyperparameter)
lenz set-objectives --objectives '{"loss":"min","flops":"min"}'
lenz set-constraints --constraints '[{"metric":"flops","upper":700}]'
\end{lstlisting}
Because all outcomes are stored independently, they are not tied to a specific role in the optimization specification.
We will demonstrate this flexibility in an example moving from constrained single-objective to multi-objective Pareto-front optimization without restarting in \cref{sec:exp:flexibility}.

\item \highlight{probe} (\texttt{diagnostics}, \texttt{predict}, \texttt{score}, \texttt{incumbent}, \texttt{pareto}, \texttt{trials}, \texttt{status}).
These read-only commands expose surrogate diagnostics, predictions, acquisition scores, incumbents, the current Pareto front, and trial history to inform the agent's decisions or further deliberation.
\begin{lstlisting}[style=lenz]
lenz diagnostics  # outputs CVR2, all the GP hyperparameters, as well as sensitivities
lenz predict --configs '[{...}]'  # outputs posterior estimates + probability of feasibility
lenz score --configs '[{...}]' --acqf logei  # scores the configs according to the acq. function
lenz incumbent  # returns the current incumbent
lenz pareto  # returns the current pareto front
lenz trials  # returns the full list of conducted experiments
lenz status  # reports the current study and pending trials
\end{lstlisting}
The \texttt{diagnostics} command helps the agent assess \emph{when to trust} the surrogate by returning cross-validated $R^2$ and sensitivity estimates for every dimension, signals that BO practitioners also use to diagnose optimization loops.
\end{enumerate}

To make \lenz output easy to parse, every command returns JSON.
\Cref{app:lenz-ref} documents the concrete output for each command.

\paragraph{Surrogate and acquisition details.}
Under the hood, \lenz uses GP surrogates with input normalization and output standardization, following standard BO practice~\citep{balandat2020botorch}.
The backend can be extended with other model classes, such as neural network-based models~\citep{li2024study,brunzema2025bayesian}, allowing \sara to perform explicit model selection.
For constrained optimization, \lenz initially optimizes the probability of feasibility and switches to constrained logEI after finding a feasible solution~\citep{gardner2014bayesian,gelbart2014bayesian,ament2023unexpected}.
For multi-objective optimization, we include acquisition functions based on expected hypervolume improvement for both the noisy and deterministic cases~\citep{daulton2020differentiable,daulton2021parallel}.
For batch calls to \texttt{suggest}, \lenz uses fantasization to ensure batch diversity.
As stated above, for all acquisition functions the optimization bounds can be adjusted by the agent at any point to focus on promising regions.

\subsection{\sara: the surrogate-augmented autoresearch agent}
\label{sec:sara}

\sara is the reasoning agent that drives the optimization and interacts with \lenz.
Her behavior is governed by the underlying LLM, a system prompt defining her as a surrogate-augmented autoresearch and BO agent, and the reference documentation for \lenz.
We provide the complete \texttt{SYSTEM.md} and \texttt{LENZ\_REF.md} in \cref{app:system-prompt,app:lenz-prompt}, respectively.
To have full control over the system prompt, we instantiate \sara in the harness pi\footnote{pi (\url{https://pi.dev/}) is available under the MIT license.} as
\begin{lstlisting}[style=sara]
pi --tools read,bash --system-prompt "$(cat SYSTEM.md)" --append-system-prompt "$(cat LENZ_REF.md)"
\end{lstlisting}

Beyond the choice of LLM, we found that the design of the system prompt does affect optimization quality.
We provide the full prompt in \cref{app:system-prompt} and summarize its most important directives here in \cref{tab:directives}.
\begin{table}[t]
\begin{threeparttable}
\caption{Core directives in \sara's system prompt and their motivation (full system prompt in \cref{app:system-prompt}).}
\label{tab:directives}
\small
\begin{tabularx}{\linewidth}{@{}p{0.55\linewidth} X@{}}
\toprule
\textbf{Directive} & \textbf{Motivation} \\
\midrule

\textbf{Agent-led, surrogate-assisted posture.} \sara owns the optimization decisions, including framing the problem, deriving priors, and choosing candidates, while treating \lenz as an instrument for calibrated search, not an autopilot.
  & We observed that without explicit ownership, the agent can become lazy and defer all decisions to the surrogate, reducing the loop to vanilla BO. \\[6pt]

\textbf{Sequential reasoning per trial.}\tnote{1}\ Each suggest$\,\to\,$evaluate$\,\to\,$observe cycle is framed as a new reasoning step; the agent must interpret the result before running the next candidate.
  & Similar to above, without per-trial reflection the agent can revert to running a full bash loop without any intermediate reasoning. \\[6pt]

\textbf{Adaptive opening strategy.} The first evaluations are chosen based on signal strength: a concrete configuration when strong priors exist, a bounded region when only a region is trusted, and space-filling points when no prior is available.
  & A fixed opening (\eg\ always Sobol) wastes budget when strong beliefs exist. We observed that \sara tended to warm up the surrogate with Sobol samples before testing her own hypothesis. \\[6pt]

\textbf{Explicit anti-patterns.} We enumerate specific failure modes: deferring to an untrustworthy posterior, recording predictions instead of observations, and discarding context signals.
  & Each failure mode was observed in early experiments, though only rarely; explicit prohibition was effective in preventing these behaviors. \\[6pt]

\textbf{Reasoning visibility.} Before every submission the agent must state what it believes and what the next evaluation should learn.
  & Explicit verbalization anchors the decision and enables post-hoc auditing. \\[6pt]

\bottomrule
\end{tabularx}
\begin{tablenotes}
\footnotesize
\item[1] This directive mitigates but does not eliminate the behavior.
Models trained as coding agents retain a turn-economy prior and, given \texttt{bash}, still issue batched and looped \texttt{suggest} calls---trading surrogate-update frequency for fewer turns.
Fully enforcing one reasoning step per evaluation would require harness-level constraints in \lenz rather than prompting alone.
\end{tablenotes}
\end{threeparttable}
\end{table}

\section{Experiments}
\label{sec:experiments}

We evaluate \sara and \lenz on standard synthetic BO functions, the LCBench AutoML benchmark, and a family of synthetic chemical-process optimization problems.
These experiments address the following research questions.
\begin{itemize}
    \item[\textbf{(Q1)}] Without any prior knowledge, can agentic BO recover standard BO performance and outperform pure LLM-based approaches?
    \item[\textbf{(Q2)}] Can \sara turn a natural-language problem description into a useful domain-informed prior, reflected in a better warm start or faster convergence?
     \item[\textbf{(Q3)}] How does \sara adapt its use of the Bayesian backend during a campaign, and can it reformulate the problem when requirements change?
    \item[\textbf{(Q4)}] How important are model family, capability, and reasoning level for the final performance?
\end{itemize}

\paragraph{On the experimental setup.}
Standard test functions (Ackley, Hartmann, \dots) and their optima are well-represented in LLM pretraining corpora.
To prevent LLM-based methods from directly exploiting memorized solutions, every synthetic benchmark is presented through an opaque programmatic interface: parameters are renamed, the domain is canonicalized to the unit hypercube, and optima are shifted away from their textbook locations.
We observed that without these measures, LLAMBO and \sara were able to one-shot problems such as Ackley whose optimum lies at the center of the domain.
Because \sara executes experiments autonomously and can inspect her working directory, we place each run in a sandbox that contains only the task description, a random token, and a symbolic link to the evaluation oracle; the sandbox is named by the token rather than the benchmark.
Neither the path nor the directory contents reveal which benchmark she is solving.
In all experiments, we report \emph{simple regret} (best-so-far gap to the known optimum, or raw best value for yield objectives) as a function of evaluation count, aggregated over 10 independent seeds.
All plots show the median with a $25$--$75\%$ interquartile band.
\Cref{app:exp-design} provides full setup details.

\paragraph{Baselines.}
We compare against four baselines (full details in \cref{app:baselines}):
\begin{itemize}
    \item \textbf{Sobol} quasi-random sampling, as a space-filling lower bound.
    \item \textbf{Ax}~\citep{olson2025ax} with its default strategy, representing a well-tuned classical BO policy with best practices implemented for all problem classes.
    \item \textbf{LLAMBO}~\citep{liu2024large}, which renders the trial history as a table and queries an LLM for the next candidate directly, without maintaining a calibrated surrogate model.
    \item \textbf{Centaur}~\citep{ferreira2026can}, a recent hybrid approach similar to our surrogate-augmented research agent that incorporates CMA-ES \citep{hansen2016cma} internal state into the context of an LLM autoresearcher.
\end{itemize}
When a natural-language problem description is available, it is provided to \sara, LLAMBO, and Centaur, ensuring the comparison isolates algorithmic differences rather than information asymmetry.
Unless otherwise specified, all LLM-based baselines use Claude Opus 4.8 with \emph{high} reasoning mode.

\subsection{Synthetic benchmarks}
\label{sec:exp:synthetic}

\begin{figure}[t]
  \centering
  \begin{subfigure}[b]{\textwidth}
    \centering
    \includegraphics[]{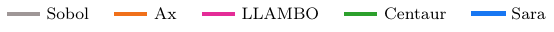}
  \end{subfigure}
  
  \centering
  \begin{subfigure}[b]{0.32\textwidth}
    \centering
    \includegraphics[width=\textwidth]{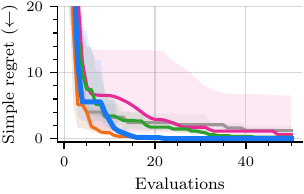}
    \subcaption{Branin (2-D)}
    \label{fig:syn:branin}
  \end{subfigure}\hfill
  \begin{subfigure}[b]{0.32\textwidth}
    \centering
    \includegraphics[width=\textwidth]{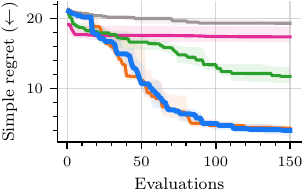}
    \subcaption{Ackley (10-D)}
    \label{fig:syn:ackley10}
  \end{subfigure}\hfill
  \begin{subfigure}[b]{0.32\textwidth}
    \centering
    \includegraphics[width=\textwidth]{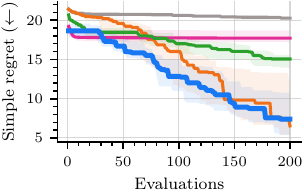}
    \subcaption{Ackley (20-D)}
    \label{fig:syn:ackley20}
  \end{subfigure}
  \vspace{5pt}
  \begin{subfigure}[b]{0.32\textwidth}
    \centering
    \includegraphics[width=\textwidth]{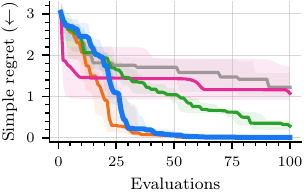}
    \subcaption{Hartmann (6-D)}
    \label{fig:syn:hartmann}
  \end{subfigure}\hfill
  \begin{subfigure}[b]{0.32\textwidth}
    \centering
    \includegraphics[width=\textwidth]{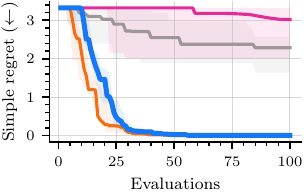}
    \subcaption{Constrained Hartmann (6-D)}
    \label{fig:syn:chartmann}
  \end{subfigure}\hfill
  \begin{subfigure}[b]{0.32\textwidth}
    \centering
    \includegraphics[width=\textwidth]{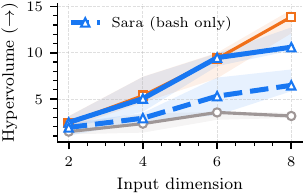}
    \subcaption{Bi-objective GP samples}
    \label{fig:syn:moo}
  \end{subfigure}
  \caption{Synthetic benchmarks. Simple regret vs.\ evaluations for the single-objective and constrained problems; final hypervolume vs.\ input dimension for the multi-objective sweep. Median over 10 seeds; shaded regions indicate the $25$--$75\%$ interquartile range.}
  \label{fig:synthetic}
\end{figure}

We first evaluate on synthetic problems where the language model receives no domain description.
This setting is adversarial for our approach in the sense that the agent has no semantic information to exploit.
The benchmark suite spans low-dimensional (Branin, 2-D), moderate-dimensional (Hartmann, 6-D), and high-dimensional (Ackley, 10-D and 20-D) unconstrained problems, as well as a constrained problem (constrained Hartmann, 6-D) and a multi-objective dimensional sweep.
The latter minimizes two independent GP sample paths and measures hypervolume relative to $(0,0)$; construction details are given in~\cref{app:gp-samples}.

Results are shown in \cref{fig:synthetic}.
\sara performs on par with Ax across the single-objective benchmark suite: she matches Ax on the low- and moderate-dimensional problems and slightly outperforms it on Ackley (20-D), where her initial space-filling strategy and adaptive search-space refinement yield faster convergence.
The multi-objective sweep shows an increasing advantage of the full system over a \texttt{bash}-only variant of \sara as dimensionality grows (more discussion on this baseline in \cref{sec:ablations} and \cref{sec:discussion:bash_only}).
The LLM-in-the-loop baselines substantially underperform on these problems, at times performing worse than random sampling.
Without recognizable structure in the trial history, their LLM-generated proposals are effectively uninformed, and these methods fail to explore the search space systematically.
In contrast, \sara uses \lenz's surrogate-backed search on these tasks, incurring no apparent penalty from the additional agentic layer.
This confirms that the framework gracefully reduces to standard BO performance when the agent has no useful domain knowledge to contribute.

\subsection{Hyperparameter optimization}
\label{sec:exp:lcbench}

\begin{figure}[t]
  \centering
  \begin{subfigure}[b]{\textwidth}
    \centering
    \includegraphics[]{assets/plots/legend.pdf}
  \end{subfigure}
  
  \centering
  \begin{subfigure}[b]{0.32\textwidth}
    \centering
    \includegraphics[width=\textwidth]{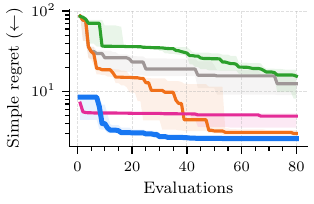}
    \subcaption{LCBench: Dionis}
    \label{fig:lc:dionis}
  \end{subfigure}\hfill
  \begin{subfigure}[b]{0.32\textwidth}
    \centering
    \includegraphics[width=\textwidth]{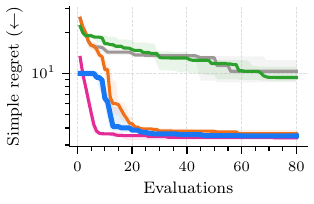}
    \subcaption{LCBench: Covertype}
    \label{fig:lc:covertype}
  \end{subfigure}\hfill
  \begin{subfigure}[b]{0.32\textwidth}
    \centering
    \includegraphics[width=\textwidth]{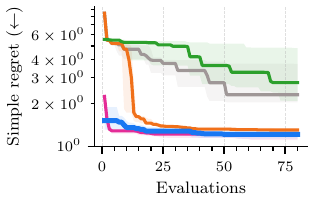}
    \subcaption{LCBench: Airlines}
    \label{fig:lc:airlines}
  \end{subfigure}
  \caption{\textit{Hyperparameter optimization on LCBench. }Simple regret (log scale) vs.\ evaluations. \sara and LLAMBO achieve strong initial performance.
  The final performance for \sara, LLAMBO, and Ax is very similar.}
  \label{fig:lcbench}
\end{figure}

We next consider a setting where domain knowledge is available and optimize neural network hyperparameters on LCBench~\citep{zimmer2021auto}.
\sara and the other LLM-based baselines receive a natural-language description of the overall training setup (see \cref{app:exp-design}).

\cref{fig:lcbench} shows a clear separation between methods that receive the problem description and those that do not.
Both \sara and LLAMBO achieve strong performance in early iterations, concentrating evaluations in the region an experienced practitioner would explore first, while Ax spends several initial evaluations discovering this region from scratch.
Centaur underperforms throughout; it first runs CMA-ES warm-up steps before querying the LLM with a fixed probability, which may limit its ability to adapt early.
On these relatively easy tasks, the prior accounts for most of the performance gap and \sara and LLAMBO achieve similar early trajectories.
The reaction-yield tasks in \cref{sec:exp:chemistry}, where the gap between a good operating regime and the optimum is larger, provide a sharper distinction between prior-only and prior-plus-surrogate approaches.

\subsection{Chemistry-informed reaction optimization}
\label{sec:exp:chemistry}

\begin{figure}[t]
  \centering
  \begin{subfigure}[b]{\textwidth}
    \centering
    \includegraphics[]{assets/plots/legend.pdf}
  \end{subfigure}\hfill
  \centering
  \begin{subfigure}[b]{0.24\textwidth}
    \centering
    \includegraphics[width=\textwidth]{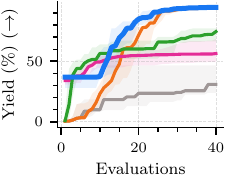}
    \subcaption{Suzuki--Miyaura}
    \label{fig:chem:suzuki}
  \end{subfigure}\hfill
  \begin{subfigure}[b]{0.24\textwidth}
    \centering
    \includegraphics[width=\textwidth]{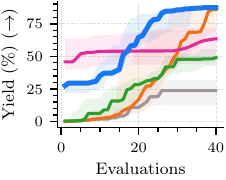}
    \subcaption{Mizoroki--Heck}
    \label{fig:chem:heck}
  \end{subfigure}\hfill
  \begin{subfigure}[b]{0.24\textwidth}
    \centering
    \includegraphics[width=\textwidth]{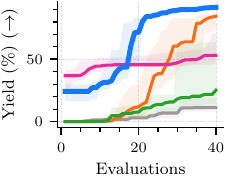}
    \subcaption{Buchwald--Hartwig}
    \label{fig:chem:buchwald}
  \end{subfigure}\hfill
  \begin{subfigure}[b]{0.24\textwidth}
    \centering
    \includegraphics[width=\textwidth]{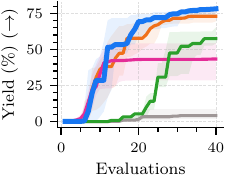}
    \subcaption{Grignard}
    \label{fig:chem:grignard}
  \end{subfigure}
  \caption{\textit{Reaction-yield optimization.}
  \sara combines a prior-informed warm-start with surrogate-based refinement, achieving fast initial convergence without sacrificing asymptotic performance.}
  \label{fig:chemistry}
\end{figure}

We construct a suite of synthetic test functions modeled on continuous-flow process optimization for named organic reactions (Suzuki--Miyaura, Mizoroki--Heck, Buchwald--Hartwig, and Grignard).
Each problem shares a six-dimensional search space consisting of temperature $T \in [0, 150]\,^\circ\mathrm{C}$, pressure $P \in [1, 10]\,\mathrm{bar}$, catalyst loading $c \in [0.1, 10]\,\mathrm{mol\%}$, residence time $t \in [1, 180]\,\mathrm{min}$, pH $\in [6, 13]$, and solvent water fraction $w \in [0, 1]$.
The optimal operating regime differs across reactions.
Full construction details are provided in \cref{app:reaction}.

\cref{fig:chemistry} shows the performance in terms of yield achieved by each method and reveals a clear separation between them.
Ax converges to near-optimal yield but does so slowly, as it must discover the productive operating regime from scratch.
LLAMBO exhibits the opposite failure mode.
The problem description encodes the search-space context and reaction name, enabling strong initial performance, but LLAMBO plateaus well below the optimum without a calibrated surrogate to refine within the identified regime.
\sara is the only method that achieves both: the language-model prior directs early evaluations to a productive region, and the Gaussian process posterior subsequently refines within that region, matching or exceeding Ax's final yield while requiring substantially fewer evaluations to reach it.

\subsection{Run-time adaptation with \sara and \lenz}
\label{sec:exp:flexibility}

We next investigate \sara's tool use during a standard optimization campaign and how it adapts when the problem specification changes mid-run.

\paragraph{Adaptive tool use.}

\begin{wrapfigure}{r}{0.44\textwidth}
    \centering
    \vspace{-1.2em}
    \includegraphics[width=\linewidth]{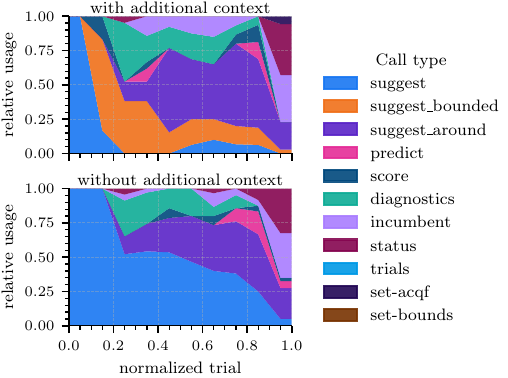}
    \caption{\textit{Tool use over an optimization campaign.}
    Relative frequency of \lenz calls on Buchwald--Hartwig with and without natural-language context, averaged over 10 seeds along normalized trial progress.}
    \label{fig:deliberation}
\end{wrapfigure}

To answer \textbf{(Q3)}, we first examine how \sara adapts her interaction with the Bayesian backend over a campaign.
\Cref{fig:deliberation} shows the relative frequency of \lenz calls over normalized trial progress on Buchwald--Hartwig, with and without the natural-language problem description.
The frequencies are averaged across 10 seeds, which smooths the individual call sequences but still reveals clear trends in tool use.
With additional context, \sara quickly moves away from generic \texttt{suggest} calls mainly used for additional Sobol coverage. 
Instead, she first requests candidates through \texttt{suggest --bounds} and then increasingly uses \texttt{suggest --around} to refine locally around the current incumbent configuration.
Without context, generic surrogate proposals remain prominent much longer, while local refinement emerges more gradually as observations accumulate.
In both settings, \texttt{diagnostics} is used primarily through the middle of the campaign, when the agent has enough data to assess the surrogate, and targeted \texttt{predict} calls appear later as the search narrows.
Near the end, \texttt{incumbent} and \texttt{status} calls become more frequent as \sara audits the final state before terminating.
Persistent mutations through \texttt{set-acqf} and \texttt{set-bounds} remain rare on this specific task.
Thus, natural-language context affects not only the initial configurations and resulting warm start but also the agent's pattern of interaction with the surrogate throughout the run.
Some full per-seed tool-use timelines, which expose the variation hidden by this average, are provided in~\cref{sec:discussion:tool_usage}.

\begin{figure}[t]
  \centering
  \includegraphics[width=\textwidth]{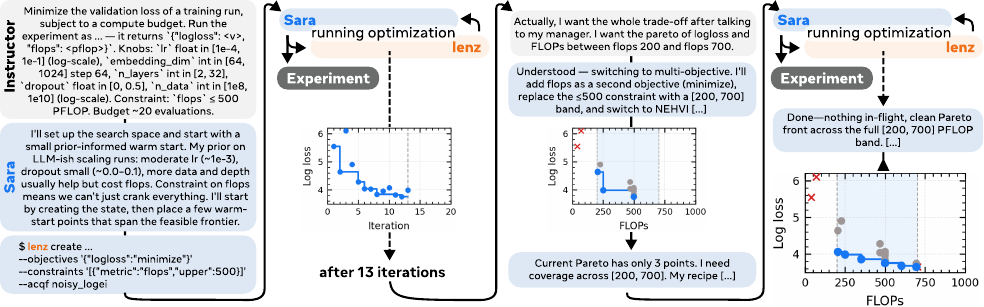}
  \caption{\textit{Mid-run reformulation on a scaling-law benchmark.} \sara first optimizes loss under a compute constraint.
  Then, after a change in requirements, \sara promotes compute to a second objective and recovers the Pareto front without restarting the campaign or manually reconfiguring the backend. \hfill\textit{(Actual data and excerpt from a trace)}
  }
  \label{fig:flexibility}
\end{figure}

\paragraph{Mid-run problem reformulation.}
Beyond adapting its tool use within a fixed problem, \sara can reconfigure the optimization specification when the requirements themselves change.
We demonstrate this capability on a synthetic neural scaling-law study in which \sara initially minimizes negative log-likelihood subject to a hard compute-budget constraint (\cf~\cref{app:scaling}; \cref{fig:flexibility} shows an excerpt from the trace).
Partway through the campaign, the instructor changes the task from finding the best model under this constraint to mapping the full loss--compute trade-off within new constraints.
In response, \sara promotes compute from a constraint to a second objective, revises the remaining constraints, and switches to a hypervolume-based acquisition function.
Because \lenz maintains the raw trial log as its single source of truth, all evaluations from the constrained phase remain valid observations under the new multi-objective specification.
The reformulation is therefore carried out directly from the natural-language instruction, without restarting the campaign, discarding data, or manually reconfiguring the backend.
This example illustrates how agentic BO can adapt not only its search strategy but also the problem being optimized as requirements evolve.

\subsection{Ablations}
\label{sec:ablations}

We next ablate various aspects of agentic BO and our specific instantiation in \sara and \lenz.

\begin{figure}[t]
  \centering
  \begin{subfigure}[b]{\textwidth}
    \centering
    \includegraphics[]{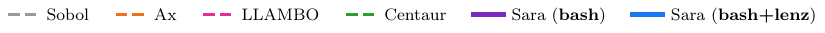}
  \end{subfigure}\hfill
  \begin{subfigure}[b]{0.24\textwidth}
    \centering
    \includegraphics[width=\textwidth]{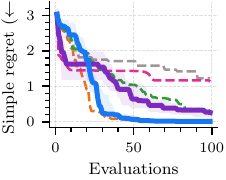}
    \subcaption{Hartmann (6-D)}
    \label{fig:surrogate-ablation:hartmann}
  \end{subfigure}\hfill
  \begin{subfigure}[b]{0.24\textwidth}
    \centering
    \includegraphics[width=\textwidth]{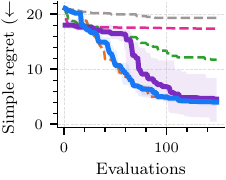}
    \subcaption{Ackley (10-D)}
    \label{fig:surrogate-ablation:ackley}
  \end{subfigure}\hfill
  \begin{subfigure}[b]{0.24\textwidth}
    \centering
    \includegraphics[width=\textwidth]{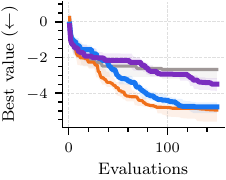}
    \subcaption{GP sample (12-D)}
    \label{fig:surrogate-ablation:gp12}
  \end{subfigure}\hfill
  \begin{subfigure}[b]{0.24\textwidth}
    \centering
    \includegraphics[width=\textwidth]{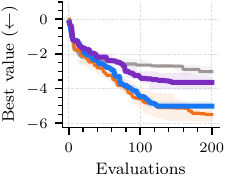}
    \subcaption{GP sample (16-D)}
    \label{fig:surrogate-ablation:gp16}
  \end{subfigure}
  \caption{\textit{Value of the surrogate backend.}
  The same \sara agent is evaluated with the full \lenz backend and with \texttt{bash} only.
  The surrogate improves reliability on established synthetic benchmarks and provides an increasingly clear advantage on high-dimensional GP sample paths that cannot be recognized from pretraining.}
  \label{fig:surrogate-ablation}
\end{figure}

\paragraph{Value of the surrogate backend.}
To isolate the contribution of \lenz's surrogate and acquisition machinery, we compare the full system against the same agent and system prompt with access to \texttt{bash} only (\cref{fig:surrogate-ablation}); \lenz remains available for bookkeeping, but its predictions, diagnostics, scoring, and proposals are hidden.
On Hartmann and Ackley, the full system converges more reliably.
The \texttt{bash}-only trajectories, however, exhibit distinct ``switching points'' after which convergence accelerates.
Inspection of the traces reveals that the LLM recognizes the benchmark from its observed structure despite the shifted optimum and renamed parameters, and subsequently exploits this knowledge.
We do not observe such explicit recognition in the full-system traces, possibly because the agent instead focuses its deliberation on steering the search through~\lenz.
We therefore additionally benchmark on random GP sample paths which, by construction, are absent from pretraining data.
On the 12-D and 16-D GP samples, the advantage of the surrogate is clear: \lenz enables systematic exploration when neither semantic priors nor recognizable benchmark structure are available.
The multi-objective results in \cref{fig:syn:moo} show the trend.
The \texttt{bash}-only agent nevertheless exhibits capable adaptive behavior, progressing from space-filling exploration to strategies such as coordinate descent, local quadratic surrogate modeling, and occasionally custom BO loops built in pure Python.
We provide the complete GP dimensional sweep, construction details, further experiments, and a longer behavioral analysis in~\cref{sec:discussion:bash_only}.
More broadly, these findings highlight an open challenge for the BO and autoresearch communities: how to evaluate LLM-based optimizers reproducibly and fairly when established benchmark functions may be recognizable from pretraining.

\paragraph{Value of natural-language priors \textbf{(Q2)}.}

\begin{figure}[t]
  \centering
      \begin{subfigure}[b]{0.66\textwidth}
    \centering
    \includegraphics[]{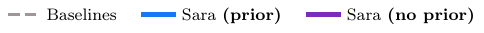}
  \end{subfigure}\hfill
  \begin{subfigure}[b]{0.3\textwidth}
    \centering
    \phantom{\includegraphics[width=\textwidth]{assets/plots/legend_prior.pdf}}
  \end{subfigure}
  \begin{subfigure}[b]{0.32\textwidth}
    \centering
    \includegraphics[width=\textwidth]{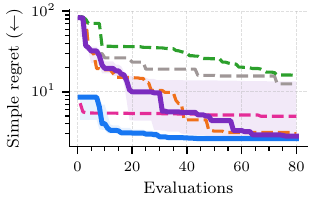}
    \subcaption{LCBench: Dionis}
    \label{fig:abl:prior:lcbench}
  \end{subfigure}\hfill
  \begin{subfigure}[b]{0.32\textwidth}
    \centering
    \includegraphics[width=\textwidth]{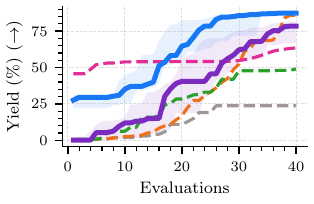}
    \subcaption{Mizoroki--Heck}
    \label{fig:abl:prior:heck}
  \end{subfigure}\hfill
  \begin{subfigure}[b]{0.32\textwidth}
    \centering
    \includegraphics[width=\textwidth]{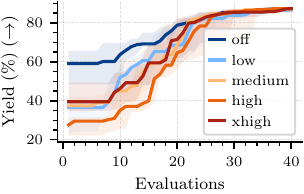}
    \subcaption{Mizoroki--Heck (reasoning level)}
    \label{fig:abl:prior:reasoning}
  \end{subfigure}
  \caption{\textit{Prior ablation.} \sara\ with vs.\ without the natural-language problem description. The prior provides a substantial reduction in early regret and, on structured problems, a better final solution. Without the prior, the surrogate eventually identifies the productive region, but requires additional evaluations to do so.}
  \label{fig:ablation-prior}
\end{figure}

We ablate the natural-language context by running \sara with and without the problem description on the same tasks in \cref{fig:ablation-prior}.
The effect is consistent with the main results.
On problems where the description encodes genuine structural knowledge, such as parameter names and experimental details, the prior yields substantially lower early regret and better final performance.
Without the prior, \sara still converges at approximately the same rate as Ax (orange dashed line).
We did not test LLAMBO and Centaur without a prior because \cref{sec:exp:synthetic} already compares these methods in the no-prior setting.

\paragraph{Effect of reasoning level \textbf{(Q4)}.}
We benchmark the effort levels \emph{off}, \emph{low}, \emph{medium}, and \emph{high} on one no-prior task, Ackley (10-D), and two prior-informed tasks, LCBench (Dionis) and Mizoroki--Heck.
The full results in \cref{app:influence_reasoning} show little effect from reasoning budget, with one exception: on Mizoroki--Heck, \emph{off} performs better in the warm-start phase than the other modes.
Even with reasoning effort set to \emph{off}, the system prompt requires \sara to state a rationale for the next submission, so the agent still produces explicit reasoning.
On this task, additional reasoning can produce a worse warm start.
For a longer discussion, we refer to \cref{app:influence_reasoning}.

\begin{figure}[t]
  \centering
    \begin{subfigure}[b]{\textwidth}
    \centering
    \includegraphics[]{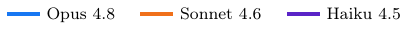}
  \end{subfigure}\hfill
  \begin{subfigure}[b]{0.32\textwidth}
    \centering
    \includegraphics[width=\textwidth]{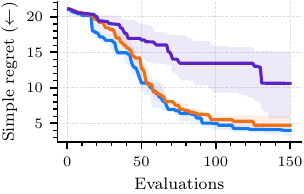}
    \subcaption{Ackley (10-D)}
    \label{fig:abl:capability:ackley}
  \end{subfigure}\hfill
  \begin{subfigure}[b]{0.32\textwidth}
    \centering
    \includegraphics[width=\textwidth]{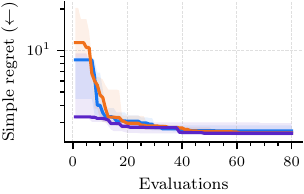}
    \subcaption{LCBench: Dionis}
    \label{fig:abl:capability:lcbench}
  \end{subfigure}\hfill
  \begin{subfigure}[b]{0.32\textwidth}
    \centering
    \includegraphics[width=\textwidth]{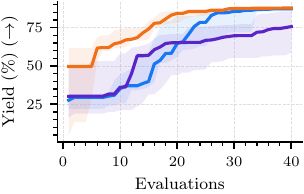}
    \subcaption{Mizoroki--Heck}
    \label{fig:abl:capability:heck}
  \end{subfigure}
  \caption{\textit{Model capability ablation.} The \sara/\lenz system with different backing LLMs, ranging from strong reasoning models to weaker alternatives.
  Effective tool use requires a minimum level of model capability.}
  \label{fig:ablation-capability}
\end{figure}

\paragraph{Effect of model capability \textbf{(Q4)}.}
We fix the \lenz backend and model instructions, varying only the LLM: Opus~4.8, Sonnet~4.6, and Haiku~4.5.
\cref{fig:ablation-capability} reveals that sensitivity to model capability is task-dependent.
On Ackley (10-D), where optimization depends heavily on competent use of \lenz, Haiku underperforms the stronger models.
On the hyperparameter and yield tasks, the performance gap across models is smaller.
This suggests that model capability matters primarily for effective and efficient use of the \lenz CLI, whereas warm-start quality is less sensitive.

\begin{figure}[t]
  \centering
  \begin{subfigure}[b]{\textwidth}
    \centering
    \includegraphics[]{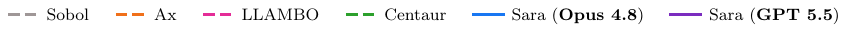}
  \end{subfigure}\hfill
  \centering
  \begin{subfigure}[b]{0.32\textwidth}
    \centering
    \includegraphics[width=\textwidth]{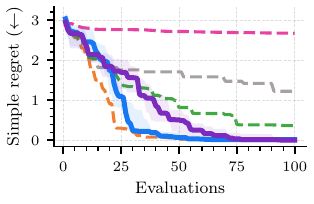}
    \subcaption{Hartmann (6-D)}
    \label{fig:mf:hartman}
  \end{subfigure}\hfill
  \begin{subfigure}[b]{0.32\textwidth}
    \centering
    \includegraphics[width=\textwidth]{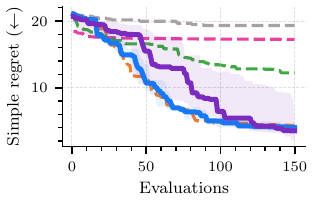}
    \subcaption{Ackley (10-D)}
    \label{fig:mf:ackley10}
  \end{subfigure}\hfill
  \begin{subfigure}[b]{0.32\textwidth}
    \centering
    \includegraphics[width=\textwidth]{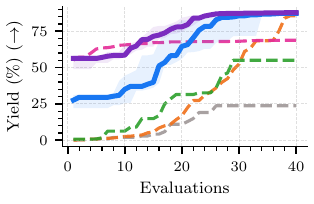}
    \subcaption{Mizoroki--Heck}
    \label{fig:mf:heck}
  \end{subfigure}
  \vspace{5pt}
  \begin{subfigure}[b]{0.32\textwidth}
    \centering
    \includegraphics[width=\textwidth]{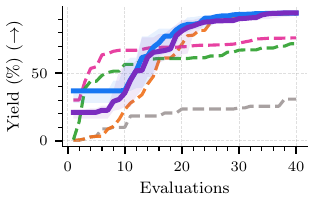}
    \subcaption{Suzuki--Miyaura}
    \label{fig:mf:suzuki}
  \end{subfigure}\hfill
  \begin{subfigure}[b]{0.32\textwidth}
    \centering
    \includegraphics[width=\textwidth]{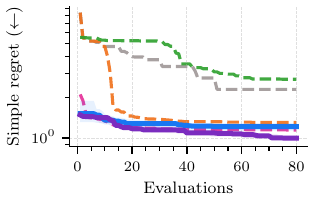}
    \subcaption{LCBench: Airlines}
    \label{fig:mf:airlines}
  \end{subfigure}\hfill
  \begin{subfigure}[b]{0.32\textwidth}
    \centering
    \includegraphics[width=\textwidth]{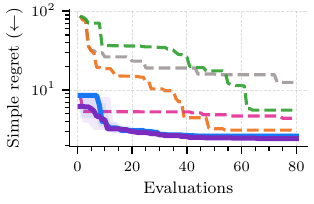}
    \subcaption{LCBench: Dionis}
    \label{fig:mf:dionis}
  \end{subfigure}
  \caption{\textit{Model family ablation.} The \sara/\lenz loop backed by models from two different LLM families.
  Both model families perform strongly relative to the baselines.}
  \label{fig:model_family}
\end{figure}
\begin{table}[t]
\centering
\definecolor{sarablue}{HTML}{1877F2}
\setlength{\aboverulesep}{0pt}
\setlength{\belowrulesep}{0pt}
\renewcommand{\arraystretch}{1.25}
\caption{Final simple regret (median over seeds; $q_{25}$ and $q_{75}$ as sub-/superscripts). Lower is better. The best median in each row and methods not significantly different from it under a two-sided Mann--Whitney test ($p\geq 0.05$) are in \textbf{bold}.}
\label{tab:gpt_vs_opus}
\resizebox{\textwidth}{!}{%
\small
\begin{tabular}{l cccc @{\hspace{1.2em}} >{\columncolor{sarablue!10}}c >{\columncolor{sarablue!10}}c}
\toprule
 & \multicolumn{4}{c}{Baselines} & \multicolumn{2}{c}{\highlight{\sara\ (ours)}} \\
\cmidrule(lr){2-5} \cmidrule(lr){6-7}
Problem & Sobol & Ax & LLAMBO & Centaur & Opus~4.8 & GPT~5.5 \\
\midrule
Branin & $1.2241^{1.9899}_{0.9120}$ & $0.0006^{0.0010}_{0.0005}$ & $0.6140^{6.5453}_{0.2023}$ & $0.2328^{0.3492}_{0.0471}$ & $0.0020^{0.0107}_{0.0006}$ & $\mathbf{0.0000}^{0.0000}_{0.0000}$ \\
Hartmann-6 & $1.2235^{1.5718}_{1.0220}$ & $\mathbf{0.0116}^{0.0435}_{0.0031}$ & $1.0756^{1.7324}_{0.9110}$ & $0.2848^{0.4196}_{0.1779}$ & $\mathbf{0.0114}^{0.0472}_{0.0025}$ & $\mathbf{0.0038}^{0.0181}_{0.0006}$ \\
Constr.\ Hartmann-6 & $2.2844^{2.5587}_{1.6398}$ & $\mathbf{0.0118}^{0.0759}_{0.0022}$ & $3.0216^{3.3222}_{1.9934}$ & -- & $\mathbf{0.0039}^{0.0218}_{0.0010}$ & $\mathbf{0.0090}^{0.0457}_{0.0005}$ \\
Ackley-10 & $19.3208^{19.6592}_{19.0833}$ & $\mathbf{4.3349}^{4.5010}_{4.1948}$ & $17.3726^{18.8270}_{16.5854}$ & $11.7241^{13.1286}_{10.7962}$ & $\mathbf{4.0033}^{4.4798}_{3.2723}$ & $\mathbf{3.5395}^{4.5155}_{2.1278}$ \\
Ackley-20 & $20.2772^{20.5228}_{20.0543}$ & $6.5389^{11.9659}_{5.3032}$ & $17.7163^{18.7241}_{17.2203}$ & $15.0561^{15.5771}_{13.9672}$ & $7.3978^{10.7848}_{5.3599}$ & $\mathbf{0.7797}^{3.9116}_{0.0055}$ \\
\midrule
Suzuki & $64.1551^{71.0712}_{48.3976}$ & $\mathbf{1.3062}^{1.9016}_{0.8983}$ & $38.6083^{45.7485}_{26.5239}$ & $20.4683^{28.4634}_{17.7271}$ & $\mathbf{0.4564}^{1.0715}_{0.3115}$ & $\mathbf{0.3365}^{2.2893}_{0.1588}$ \\
Heck & $64.2884^{75.0154}_{56.1902}$ & $2.2505^{18.6613}_{1.9684}$ & $24.5770^{42.6077}_{16.2859}$ & $39.0392^{45.1923}_{32.4834}$ & $0.7301^{2.2925}_{0.3864}$ & $\mathbf{0.1975}^{0.4127}_{0.1357}$ \\
Buchwald & $80.6547^{83.9159}_{72.9054}$ & $7.4335^{18.1997}_{5.2334}$ & $38.7766^{45.5889}_{21.4640}$ & $66.8719^{77.9784}_{28.0592}$ & $\mathbf{0.3391}^{1.6288}_{0.2327}$ & $\mathbf{0.1564}^{4.1827}_{0.0595}$ \\
Grignard & $75.8200^{78.1048}_{71.2676}$ & $7.0111^{10.5178}_{2.4048}$ & $36.6279^{51.3405}_{25.6798}$ & $22.5169^{25.9783}_{14.7961}$ & $1.8336^{6.1981}_{0.7324}$ & $\mathbf{0.0672}^{0.1536}_{0.0000}$ \\
LCBench Airlines & $2.2888^{3.7445}_{2.0277}$ & $1.3032^{1.3312}_{1.2732}$ & $1.2149^{1.2505}_{1.1073}$ & $2.7764^{4.8164}_{2.0743}$ & $1.2140^{1.2884}_{1.1335}$ & $\mathbf{1.0007}^{1.0440}_{0.9646}$ \\
LCBench Covertype & $10.4014^{12.5400}_{8.6155}$ & $\mathbf{3.6258}^{3.8541}_{3.4690}$ & $\mathbf{3.3835}^{3.5729}_{3.2524}$ & $9.3576^{11.1695}_{8.8375}$ & $\mathbf{3.4976}^{3.5885}_{3.4229}$ & $\mathbf{3.4077}^{3.4948}_{3.1035}$ \\
LCBench Dionis & $12.5219^{16.0751}_{9.0688}$ & $3.0078^{3.3239}_{2.8663}$ & $4.9541^{5.3412}_{3.5565}$ & $15.5670^{18.6681}_{8.1948}$ & $\mathbf{2.6046}^{2.8546}_{2.4723}$ & $\mathbf{2.4248}^{2.4550}_{2.3042}$ \\
\bottomrule
\end{tabular}%
}
\end{table}

\paragraph{Effect of model family \textbf{(Q4)}.}
All the experiments so far were conducted with the Claude family of models.
To test whether \sara's performance depends on one model family, we compare Opus~4.8 with GPT~5.5 in \cref{fig:model_family}.
\sara with GPT~5.5 also performs strongly relative to the baselines across the benchmark suite.
GPT~5.5 tends to perform better on prior-informed benchmarks (\cref{fig:mf:heck} to \cref{fig:mf:dionis}), whereas Opus~4.8 converges faster on the no-prior benchmarks.
A full summary of the final performance across baselines and benchmarks is provided in \cref{tab:gpt_vs_opus}.
This stronger performance does, however, come at the cost of a higher total token count (\cf~\cref{app:token_usage}).
Note that we use the same system prompt for both model families.
Tuning the system prompt to a specific model or model family could further improve performance.

\section{Limitations}
\label{sec:limitations}

Agentic Bayesian optimization and surrogate-augmented autoresearch offer a fundamentally different approach to optimizing expensive-to-evaluate functions.
Still, our current instantiation has limitations that point to concrete directions for future work.

\paragraph{Structured search spaces.}
\sara and \lenz assume a well-defined parameter space, \ie, bounded reals, categoricals, and ordinals, over which a GP surrogate is straightforward to specify.
Open-ended settings where optimization is conducted over code diffs or other complex design artifacts fall outside this scope.
Because the agent-in-the-loop architecture is agnostic to the backend, extending to such spaces is a matter of encoding and surrogate design rather than a change to the loop itself.
Na\"ively encoding each diff as a categorical variable is possible but scales poorly and discards structural similarities between candidates; specifying or learning a representation that preserves such structure, potentially with the agent's help, is a compelling direction for future work.

\paragraph{Sensitivity to prompt and interface design.}
\sara's decisions are shaped by the system prompt, the tool descriptions of \lenz, and the provided problem statement.
We found that the wording of system-prompt directives noticeably influences the agent's behavior.
In \cref{tab:directives}, we list core directives that we found to improve interaction with \lenz and make the behavior more consistent across seeds.
Defining robust ways for agents to call tools and interact with surrogates therefore remains especially important for the broader direction of agentic BO.

\paragraph{Non-determinism.}
Because an LLM drives the decision loop, repeated runs on an identical problem will generally follow different trajectories (\cref{fig:tool-usage-timelines}).
This introduces failure modes absent from classical BO.
We additionally observe a form of \emph{tool-use inertia}: whichever calling pattern the agent adopts early, \eg relying heavily on \texttt{suggest} versus \texttt{predict} or \texttt{score}, tends to persist throughout the run, reducing the diversity of strategies explored.
Maintaining tool-call diversity, whether through prompting or architectural interventions, is an important direction for future work.
We report distributions over seeds to surface these effects.

\section{Conclusion and Outlook}
\label{sec:conclusion}

We introduced \highlight{agentic Bayesian optimization}, a setting in which an LLM agent drives the optimization loop while a Bayesian backend supplies principled posterior uncertainty estimates through a probabilistic surrogate as well as access to popular acquisition functions.
We instantiated this paradigm in \sara and \lenz.
\lenz is a modular BO backend whose raw trial log is the single source of truth; reconfiguration is cheap, lossless, and designed from the ground up for agentic use.
\sara is the reasoning agent that drives the loop: she frames the problem from natural-language context, derives priors, decides what to evaluate, and steers the search by editing bounds, acquisition functions, objectives, and constraints mid-campaign.

Without semantic context, \sara matches standard BO on synthetic benchmarks and outperforms LLM-based optimizers, preserving systematic search when no useful prior is available.
On hyperparameter and reaction-yield benchmarks, natural-language descriptions guide the search toward well-performing regions, while the Bayesian backend supports continued refinement.
When requirements change mid-run, \sara is able to reconfigure the full optimization specification through \lenz.

The framework opens two complementary directions for extension: broadening the backend’s modeling capabilities and improving the agent’s deliberation over them.
On the backend side, extending \lenz to search spaces where configurations are structured objects, \eg programs or code diffs, requires richer surrogate classes and structured-input kernels.
More generally, we believe that advances in BO methodology will also directly benefit agentic BO.
On the agent side, more robust tool use or fine-tuned BO agents could further improve downstream performance.
More broadly, agentic BO will be most suited to campaigns in which evaluations are expensive, domain knowledge is available, and requirements may evolve during the search.
We see agentic BO as a path toward autonomous optimization campaigns that can be steered conversationally while combining the sample efficiency and systematic exploration of BO with the vast prior knowledge and sequential decision-making capabilities of LLMs.

\par\vspace{5em}
\begin{center}
    \begin{minipage}[c]{0.6\linewidth}
        \centering
        \begin{minipage}[c]{0.28\linewidth}
            \includegraphics[width=\linewidth]{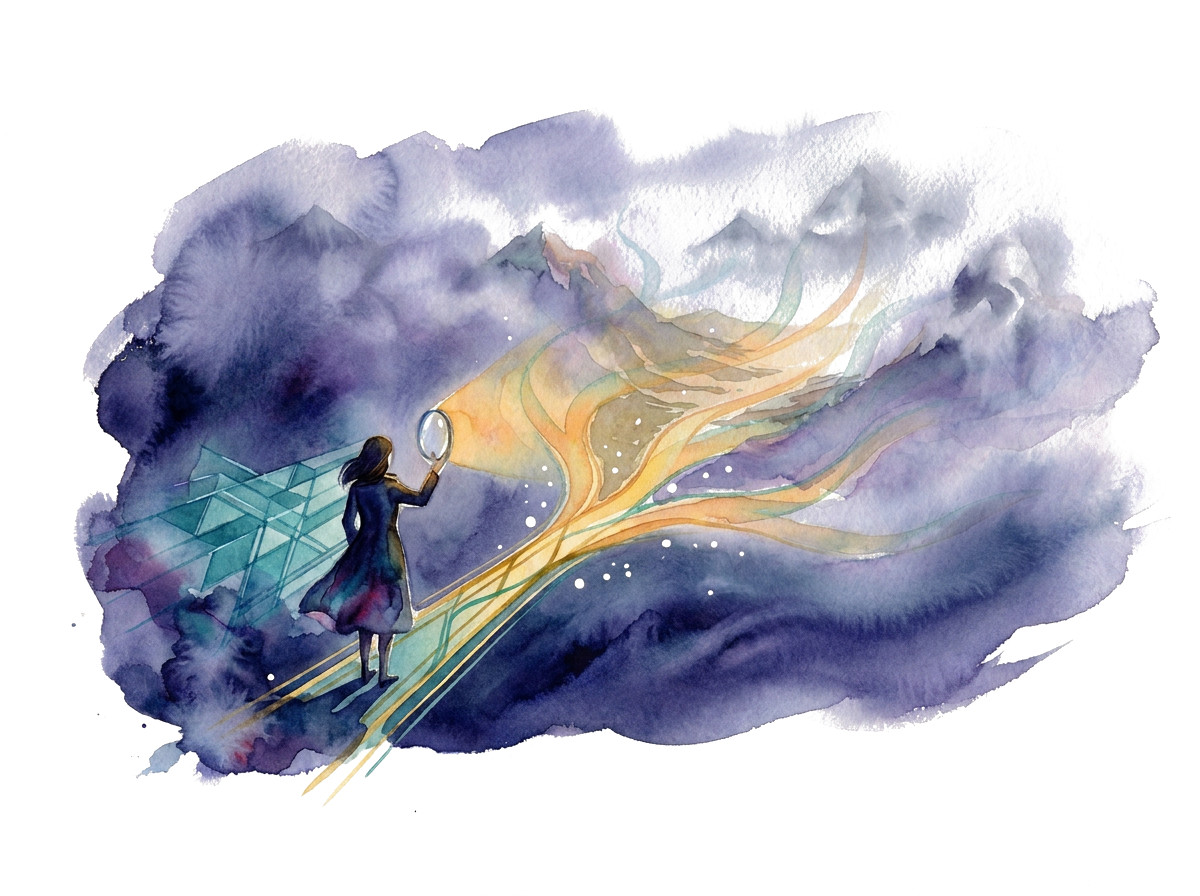}
        \end{minipage}%
        \hspace{1em}%
        \begin{minipage}[c]{0.62\linewidth}
            \scriptsize
            \emph{\sara{} uses \lenz{} to find the best path}\\
            \emph{to the peak in an uncertain environment.}
        \end{minipage}%
    \end{minipage}%
\end{center}

\clearpage
\newpage
\section*{Acknowledgments}
We thank Maximilian Balandat, David Eriksson, and Eytan Bakshy for insightful discussions that helped shape the idea of agentic Bayesian optimization.
P.B. also thanks James Harrison, James Requeima, and Jasper Snoek for early discussions on the use of LLMs in Bayesian optimization.

\bibliographystyle{assets/plainnat}
\bibliography{paper}

\clearpage
\newpage
\beginappendix

\section{Experimental design}
\label{app:exp-design}

This appendix details the benchmark suite, the agent and backend configuration, the GP sample paths, the reaction-yield family, and the neural scaling-law study used for the mid-run reformulation of \cref{sec:exp:flexibility}.
We follow the evaluation protocol of \cref{sec:experiments} with every method running under the same evaluation budget, and we report simple regret---or the raw best value for the yield objectives---against the number of evaluations.

\paragraph{Benchmark suite and budgets.}
\Cref{tab:benchmarks} lists the problems, their dimensionality, and the per-problem evaluation budget.
The suite spans low- to high-dimensional synthetic functions, constrained and multi-objective variants, the LCBench hyperparameter-optimization surrogates~\citep{zimmer2021auto}, and the reaction-yield family described below.
Ax uses a short Sobol initialization of five trials.
\sara and LLAMBO begin proposing from the first trial, with \sara choosing her own initialization strategy through \lenz.
Centaur runs CMA-ES alone for the first ten trials before the LLM begins to intervene as described in \citet{ferreira2026can}.

\begin{table}[h]
\centering
\caption{Benchmark suite and evaluation budgets. Synthetic problems are optimized in the normalized unit cube; the reaction-yield family shares a single 6-D operating space across four reactions.}
\label{tab:benchmarks}
\small
\begin{tabular}{lcc}
\toprule
\textbf{Benchmark} & \textbf{Dimensionality} &\textbf{ Budget (evaluations)} \\
\midrule
Branin                     & 2-D                       & 50  \\
Hartmann                   & 6-D                       & 100 \\
Ackley                     & 10-D                      & 150 \\
Ackley                     & 20-D                      & 200 \\
Constrained Hartmann       & 6-D        & 100 \\
Bi-objective GP samples   & 2-, 4-, 6-, 8-D          & 40, 60, 80, 120 \\
LCBench (HPO surrogate)    & 7-D                       & 80  \\
Reaction yield (all four variants) & 6-D                      & 40  \\
\bottomrule
\end{tabular}
\end{table}

\paragraph{Agent and backend.}
Unless noted, \sara is backed by Claude Opus~4.8; the model-capability ablation (\cref{sec:ablations}) additionally reports Sonnet~4.6 and Haiku~4.5, and the reasoning ablation sweeps the effort levels \emph{off}, \emph{low}, \emph{medium}, and \emph{high}.
The \lenz backend fits a GP surrogate with a Mat\'ern kernel, normalizes inputs to the unit cube, and standardizes outcomes.
By default, it uses variants of log expected improvement for single-objective and constrained problems and the corresponding hypervolume-based acquisition for multi-objective problems.
Outcome constraints are handled through the posterior probability of feasibility as $\mathbb{P}(c(\vx) \le 0)$ (\cf \cref{app:lenz-ref}).
\sara receives the same evaluation budget as the baselines.

\paragraph{Preventing benchmark recognition.}
Because textbook benchmark functions and their optima are well represented in pretraining corpora, each synthetic optimum is relocated per seed by up to a quarter of every dimension's range, the domain is canonicalized to the unit cube, and the parameters are renamed.
Each \sara run additionally executes inside a sandbox that contains only the task description and a symbolic link to the evaluation oracle and is named by a random token, so neither the path nor the directory contents reveal which benchmark is being solved (\cf~\cref{sec:experiments,sec:discussion}).

\paragraph{Natural-language priors.}
Where a description is available---for the LCBench and reaction-yield tasks---the \emph{same} prior text is provided to \sara, LLAMBO, and Centaur, so that any difference reflects the algorithm rather than the information supplied.
Each prior conveys genuine domain knowledge (the model family and parameter scales for LCBench, the reaction class) without revealing parameter values or, for LCBench, the identity of the dataset.
The synthetic benchmarks receive no prior and thus probe the no-knowledge regime.

\begin{figure}
    \centering
    {\footnotesize
    \tcbinputlisting{
        colback=saraGrey!10,
        colframe=saraGrey!10,
        boxrule=0.0pt,
        arc=5pt,
        left=3pt, right=3pt, top=3pt, bottom=3pt,
        listing only,
        listing file=assets/lcbench.md,
        listing options={
            basicstyle=\footnotesize\ttfamily,
            columns=fullflexible,
        }
    }}
    \caption{Problem description for all LCBench benchmarks.}
    \label{fig:app:prompt_lcbench}
\end{figure}

\begin{figure}
    \centering
    {\footnotesize
    \tcbinputlisting{
        colback=saraGrey!10,
        colframe=saraGrey!10,
        boxrule=0.0pt,
        arc=5pt,
        left=3pt, right=3pt, top=3pt, bottom=3pt,
        listing only,
        listing file=assets/reactions.md,
        listing options={
            basicstyle=\footnotesize\ttfamily,
            columns=fullflexible,
        }
    }}
    \caption{\textit{Problem description for the Reaction-yield benchmarks.} The specific description changes, but all problems share the same search space.}
    \label{fig:app:prompt_yield}
\end{figure}

\subsection{GP sample-path benchmarks}
\label{app:gp-samples}

We draw approximate squared-exponential GP samples using $M=1028$ random Fourier features~\citep{rahimi2007random},
\begin{equation}
    f_i(\vx) = \sum_{m=1}^M w_m \phi_m(\vx),
    \qquad
    \phi_m(\vx) = \sqrt{\frac{2}{M}}\cos(\bm{\theta}_m^\top\vx+\tau_m),
\end{equation}
where $w_m\overset{\mathrm{iid}}{\sim}\mathcal{N}(0,1)$, $\bm{\theta}_m$ follows the kernel's spectral density, $\tau_m\sim\mathcal{U}(0,2\pi)$, and all lengthscales are $0.2$.
For the bi-objective sweep, we independently draw two such paths, minimize both, and measure hypervolume relative to $(0,0)$ in dimensions $2$, $4$, $6$, and $8$, with evaluation budgets of $40$, $60$, $80$, and $120$, respectively.

\subsection{Reaction-yield family}
\label{app:reaction}

We model continuous-flow process optimization for four named organic reactions.
The reactions share a six-dimensional operating space---temperature $T \in [0, 150]\,^\circ\mathrm{C}$, pressure $P \in [1, 10]\,\mathrm{bar}$, catalyst loading $c \in [0.1, 10]\,\mathrm{mol\%}$, residence time $t \in [1, 180]\,\mathrm{min}$, pH $\in [6, 13]$, and solvent water fraction $w \in [0, 1]$---but each favors a different operating regime.

The yield is a product of phenomenological factors, each valued in $(0, 1]$ and equal to one at its regime-specific optimum, so that the global maximum $y_{\max}$ is attained at a unique, known configuration:
\begin{equation}
\label{eq:reaction_yield}
y(T, P, c, t, \mathrm{pH}, w) \;=\; y_{\max} \cdot f_T \cdot f_P \cdot f_c \cdot f_t \cdot f_{\mathrm{pH}} \cdot f_w \cdot f_{\mathrm{int}}.
\end{equation}
Temperature and catalyst loading use asymmetric Gaussians, which capture Arrhenius kinetics and incomplete conversion below the optimum and thermal decomposition or catalyst deactivation above it:
\begin{equation}
f_T = \exp\!\Bigl(-\tfrac{1}{2}\bigl((T - \mu_T)/\sigma_T(T)\bigr)^2\Bigr), \quad
\sigma_T(T) = \begin{cases} \sigma_T^{-} & T < \mu_T \\ \sigma_T^{+} & T \geq \mu_T \end{cases}
\end{equation}
\begin{equation}
f_c = \exp\!\Bigl(-\tfrac{1}{2}\bigl((c - \mu_c)/\sigma_c(c)\bigr)^2\Bigr), \quad
\sigma_c(c) = \begin{cases} \sigma_c^{-} & c < \mu_c \\ \sigma_c^{+} & c \geq \mu_c \end{cases}
\end{equation}
Residence time is a saturating ramp normalized to one at the upper bound $t_{\max} = 180\,\mathrm{min}$; pressure, pH, and water fraction are symmetric Gaussians:
\begin{equation}
f_t = \frac{1 - \exp\bigl(-\max(t - t_0,\, 0) / \tau\bigr)}{1 - \exp\bigl(-(t_{\max} - t_0) / \tau\bigr)},
\end{equation}
\begin{equation}
f_P = \exp\!\Bigl(-\tfrac{1}{2}\bigl((P - \mu_P)/\sigma_P\bigr)^2\Bigr), \quad
f_{\mathrm{pH}} = \exp\!\Bigl(-\tfrac{1}{2}\bigl((\mathrm{pH} - \mu_{\mathrm{pH}})/\sigma_{\mathrm{pH}}\bigr)^2\Bigr), \quad
f_w = \exp\!\Bigl(-\tfrac{1}{2}\bigl((w - \mu_w)/\sigma_w\bigr)^2\Bigr).
\end{equation}
Finally, a temperature--water interaction term lets the ideal water content drift linearly with temperature:
\begin{equation}
f_{\mathrm{int}} = \exp\!\Bigl(-\tfrac{1}{2}\Bigl(\frac{w - \bigl[\mu_w + \beta\,(T - \mu_T)\bigr]}{\sigma_{\mathrm{int}}}\Bigr)^{\!2}\Bigr).
\end{equation}

We instantiate four reactions whose regime parameters reflect their qualitative process requirements (\cref{tab:reaction_regimes}): Suzuki--Miyaura coupling (warm, aqueous, mildly basic; $y_{\max} = 95\%$), the Mizoroki--Heck reaction (hot, dry; $88\%$), Buchwald--Hartwig amination (hot, anhydrous, strongly basic; $92\%$), and Grignard addition (cold, rigorously anhydrous; $80\%$).
The context string given to the agent names the reaction and its catalyst system, solvent, and base---enough chemical information to identify the favorable operating regime, but no explicit parameter values (\cf \cref{fig:app:prompt_yield}).
An optimizer without domain knowledge must instead discover each regime from function evaluations alone.

\begin{table}[t]
\centering
\caption{Optimal regimes for the reaction-yield family. Each reaction shares the same 6-D domain but differs in the location and sensitivity of its optimum. The residence-time optimum is $t_{\max} = 180\,\mathrm{min}$ for all reactions.}
\label{tab:reaction_regimes}
\small
\begin{tabular}{lcccccc}
\toprule
Reaction & $y_{\max}$ (\%) & $\mu_T$ ($^\circ$C) & $\mu_P$ (bar) & $\mu_c$ (mol\%) & $\mu_{\mathrm{pH}}$ & $\mu_w$ \\
\midrule
Suzuki--Miyaura   & 95 & 70  & 1.5 & 3.5 & 9.5  & 0.45 \\
Mizoroki--Heck    & 88 & 120 & 1.5 & 2.0 & 9.0  & 0.10 \\
Buchwald--Hartwig & 92 & 100 & 1.5 & 1.5 & 11.5 & 0.03 \\
Grignard          & 80 & 5   & 1.5 & 0.5 & 8.0  & 0.00 \\
\bottomrule
\end{tabular}
\end{table}

\Cref{fig:chemistry} shows the result on this family of benchmarks.
The natural-language context lets \sara locate each operating regime with far fewer evaluations than an uninformed optimizer.
In contrast to the other LLM-based baselines, \sara can use the BO backend to further fine-tune her initial solution.

\subsection{Neural scaling-law study}
\label{app:scaling}

The mid-run reformulation experiment of \cref{sec:exp:flexibility} uses a synthetic neural scaling-law benchmark that maps a model-and-training configuration to two outcomes, validation loss and training compute, without ever training a real model.
The search space comprises five knobs (\cref{tab:scaling_space}): learning rate~$\eta$, model width~$d$ and depth~$\ell$ (which together set the parameter count $N = 12\,\ell\,d^2$), dropout~$\rho$, and dataset size~$D$ (in tokens).
The two outcomes are the validation cross-entropy \texttt{logloss} and the \texttt{flops}, reported in PFLOP; both are to be minimized.

\begin{table}[h]
\centering
\caption{Search space for the neural scaling-law study.  Width and depth set the parameter count $N$; the dataset size is the token count $D$.}
\label{tab:scaling_space}
\small
\begin{tabular}{llll}
\toprule
\textbf{Parameter} & \textbf{Range} & \textbf{Scale} & \textbf{Type} \\
\midrule
\texttt{lr} (learning rate $\eta$)          & $[10^{-4},\,10^{-1}]$ & log    & float \\
\texttt{embedding\_dim} (width $d$)         & $[64,\,1024]$, step $64$ & linear & int   \\
\texttt{n\_layers} (depth $\ell$)           & $[2,\,32]$            & linear & int   \\
\texttt{dropout} ($\rho$)                    & $[0,\,0.5]$           & linear & float \\
\texttt{n\_data} (tokens $D$)               & $[10^{8},\,10^{10}]$  & log    & float \\
\bottomrule
\end{tabular}
\end{table}

Following a Chinchilla-style parametric form~\citep{hoffmann2022training}, the loss floor decreases with both model size and data:
\begin{equation}
\label{eq:scaling_floor}
L_0(N, D) = E + \frac{A}{N^{\alpha}} + \frac{B}{D^{\beta}},
\end{equation}
with $E = 1.69$, $A = 406$, $\alpha = 0.34$, $B = 410$, and $\beta = 0.28$.
Reaching that floor requires appropriate hyperparameters; the realized loss adds a log-quadratic penalty around a size-dependent learning-rate optimum and a mild dropout penalty:
\begin{equation}
\label{eq:scaling_loss}
\texttt{logloss} = L_0(N, D)
    + C_{\mathrm{lr}}\bigl(\log_{10}\eta - \log_{10}\eta^{\star}(N)\bigr)^2
    + C_{\mathrm{drop}}\,(\rho - \rho^{\star})^2
    + \varepsilon,
\qquad
\eta^{\star}(N) = \eta_{\mathrm{ref}}\Bigl(\frac{N_{\mathrm{ref}}}{N}\Bigr)^{s},
\end{equation}
where $C_{\mathrm{lr}} = 0.6$, $C_{\mathrm{drop}} = 1.2$, $\rho^{\star} = 0.1$, $\eta_{\mathrm{ref}} = 10^{-2}$, $N_{\mathrm{ref}} = 10^{7}$, $s = 0.12$, and $\varepsilon \sim \mathcal{N}(0, 0.01^2)$ is measurement noise.
Because the optimal learning rate $\eta^{\star}$ shrinks as the model grows, $\eta$ and model size interact.
Training compute follows the standard estimate $\texttt{flops} = 6\,ND / 10^{15}\;\mathrm{[PFLOP]}$.

In the experiment, \sara receives the search space and objectives in natural language and works on a single shared trial log across two phases.
In \textbf{Phase~1} she minimizes \texttt{logloss} subject to a hard constraint $\texttt{flops} \le 500\,\mathrm{PFLOP}$---a constrained single-objective campaign with roughly $20$ evaluations.
\lenz handles the constraint via the posterior feasibility $\mathbb{P}(\texttt{flops} \le 500)$ and optimizes the noisy constrained log-EI (\cf~\cref{app:lenz-ref}).
Partway through, the user changes the requirement to instead recover the full loss--compute Pareto front within $\texttt{flops} \in [200,\,700]\,\mathrm{PFLOP}$.
In \textbf{Phase~2} \sara promotes \texttt{flops} from a constraint to a second objective (\cmd{set-objectives}), relaxes the hard budget to the requested band (\cmd{set-constraints}), and switches to a hypervolume acquisition function via \cmd{set-acqf} (\cf~\cref{fig:flexibility}).
Because the accumulated observations remain valid under the new specification, no data is discarded; the reconfiguration requires only natural-language instructions to \sara, illustrating the flexibility that agentic BO affords over a conventional fixed-objective pipeline.

\begin{figure}[h]
    \centering
    \begin{subfigure}[b]{0.99\textwidth}
        \centering
        {\footnotesize
        \tcbinputlisting{
            colback=saraGrey!10,
            colframe=saraGrey!10,
            boxrule=0.0pt,
            arc=5pt,
            left=3pt, right=3pt, top=3pt, bottom=3pt,
            listing only,
            listing file=assets/scaling_so.md,
            listing options={
                basicstyle=\footnotesize\ttfamily,
                columns=fullflexible,
            }
        }}
        \vspace{-1em}
        \subcaption{Prompt for \textbf{Phase~1}}
      \end{subfigure}\\
    \begin{subfigure}[b]{0.99\textwidth}
        \centering
        {\footnotesize
        \tcbinputlisting{
            colback=saraGrey!10,
            colframe=saraGrey!10,
            boxrule=0.0pt,
            arc=5pt,
            left=3pt, right=3pt, top=3pt, bottom=3pt,
            listing only,
            listing file=assets/scaling_mo.md,
            listing options={
                basicstyle=\footnotesize\ttfamily,
                columns=fullflexible,
            }
        }}
        \vspace{-1em}
        \subcaption{Prompt for \textbf{Phase~2}}
      \end{subfigure}
    \caption{\textit{Prompts for the mid-run reformulation from \cref{sec:exp:flexibility}.}}
    \label{fig:app:prompt_scaling_so}
\end{figure}

\section{Details on baselines}
\label{app:baselines}

We compare against four baselines that include pseudo-random sampling, classical Bayesian optimization, and LLM-driven optimization.
Where a natural-language description is available, it is given to every LLM-driven method (\cref{app:exp-design}), so that comparisons isolate the optimization strategy rather than the information supplied or the underlying model.

\paragraph{Sobol.}
Scrambled Sobol quasi-random sampling, run for the full evaluation budget without any model.
It serves as a space-filling lower bound.

\paragraph{Ax.}
We use Ax~\citep{olson2025ax} with its default generation strategy and no custom overrides, representing a well-tuned classical BO policy with best practices implemented as defaults.
A short Sobol initialization of five trials is followed by a model-based search using a GP surrogate with a Mat\'ern kernel and a log-expected-improvement acquisition function, with the noisy hypervolume analogue used automatically on multi-objective problems.
For the constrained problem, Ax defaults to constrained log-expected improvement.

\paragraph{LLAMBO.}
Following \citet{liu2024large}, LLAMBO renders the trial history as a table and queries the LLM for the next configuration directly, without fitting a calibrated surrogate.
It proposes from the first trial, with no Sobol warm-up, aiming to leverage the full warm-start capability of LLMs.

\paragraph{Centaur.}
Following \citet{ferreira2026can}, Centaur shares the internal state of a CMA-ES optimizer with an LLM autoresearcher, and we use the same hyperparameters as the original paper.
CMA-ES is warm-started on its own for the first ten trials before the LLM begins to intervene; thereafter the LLM proposes a fraction of the evaluations (30\%).
The initial step size of CMA-ES is set to $\sigma_0 = 1/6$ on the unit cube and the remaining CMA-ES settings left at their defaults.
Centaur observes trial outcomes but, unlike \sara, is exposed to the optimizer state only through its prompt but cannot interact with the optimizer in any way; this is the closest baseline in spirit to agentic BO.

\paragraph{LLM backend.}
Unless otherwise specified, all LLM-driven baselines as well as \sara are driven by Claude Opus 4.8 (\textit{high}), so that comparisons reflect the optimization strategy rather than the capability of the model.

\section{Further experiments}
\label{app:further_experiments}

In the following, we list a few additional results and discuss some ablations in more detail.

\subsection{Influence of reasoning budget}
\label{app:influence_reasoning}

To assess how important the reasoning budget is for downstream optimization performance (\textbf{Q4}), we ablate the extended thinking effort level of the backing LLM across four settings: \emph{off}, \emph{low}, \emph{medium}, and \emph{high}.
We evaluate on three representative tasks spanning both the no-prior regime (Ackley 10-D) and the prior-informed regime (LCBench: Dionis, Mizoroki--Heck).

\begin{figure}[t]
  \centering
  \begin{subfigure}[b]{0.32\textwidth}
    \centering
    \includegraphics[width=\textwidth]{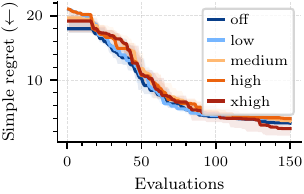}
    \subcaption{Ackley (10-D)}
    \label{fig:abl:reasoning:ackley}
  \end{subfigure}\hfill
  \begin{subfigure}[b]{0.32\textwidth}
    \centering
    \includegraphics[width=\textwidth]{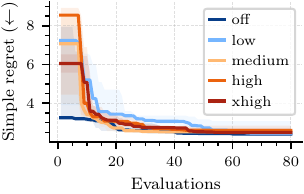}
    \subcaption{LCBench: Dionis}
    \label{fig:abl:reasoning:lcbench}
  \end{subfigure}\hfill
  \begin{subfigure}[b]{0.32\textwidth}
    \centering
    \includegraphics[width=\textwidth]{assets/plots/cmp_reasoning_heck_yield.pdf}
    \subcaption{Mizoroki--Heck}
    \label{fig:abl:reasoning:heck}
  \end{subfigure}
  \caption{\textit{Ablation of reasoning level on final performance.} We vary the extended thinking effort from \emph{off} to \emph{high} and find little effect on \sara's performance. On the Mizoroki--Heck task, \emph{off} slightly outperforms higher reasoning budgets in the warm-start phase, suggesting that additional deliberation can occasionally override a strong context-derived prior.}
  \label{fig:ablation-reasoning}
\end{figure}

\cref{fig:ablation-reasoning} reports the results.
Across all three benchmarks, reasoning level has little effect on performance.
On the synthetic Ackley function and the LCBench hyperparameter task, all effort levels converge to comparable final regret.
On the Mizoroki--Heck reaction benchmark, the \emph{off} setting outperforms higher reasoning modes in early iterations.
We hypothesize that this occurs because \sara's system prompt requires her to state a rationale before every submission, so she still produces explicit reasoning at zero reasoning budget.
On this particular task, where the natural-language prior directly identifies the productive operating regime, additional reasoning appears to occasionally lead the agent to second-guess a strong context signal and explore less promising configurations in the initial phase to also seed the GP model, resulting in a slightly worse warm start.

\subsection{Token usage per problem}

\label{app:token_usage}

\begin{figure}[t]
    \centering
    \includegraphics[width=\linewidth]{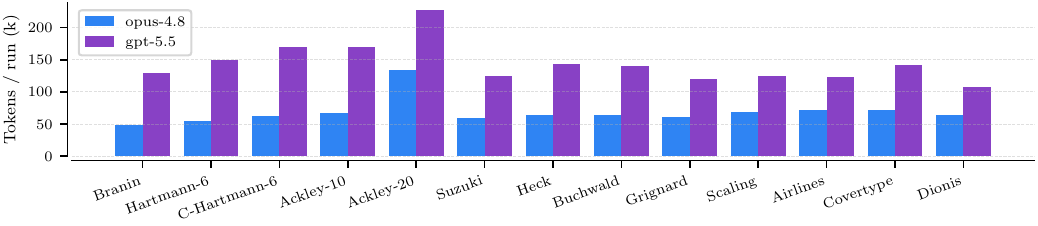}
    \caption{\textit{Token usage for the two different high-capability models used for \sara.}
    GPT~5.5 is more verbose and uses tools more frequently than Opus~4.8, resulting in higher token consumption.
    Token usage scales with evaluation budget.}
    \label{fig:token-usage}
\end{figure}

\cref{fig:token-usage} reports the total token usage per optimization run (in thousands) for the two high-capability model families used to back \sara: Opus~4.8 and GPT~5.5.
Several patterns emerge.
First, token consumption scales approximately with the evaluation budget: problems with more allowed evaluations (e.g., Ackley-20 with 200 evaluations) consume more tokens than shorter campaigns (e.g., Branin with 50 evaluations or the reaction-yield tasks with 40).
Second, GPT~5.5 is consistently more verbose than Opus~4.8 across all benchmarks, consuming roughly $1.5$--$2\times$ more tokens per run.
This increased token count reflects both more extensive reasoning traces and a higher frequency of tool calls.
GPT~5.5 tends to issue more \texttt{predict}, \texttt{score}, and \texttt{diagnostics} queries to \lenz per trial, resulting in longer interaction traces.
Furthermore, Opus~4.8 tends to stop earlier than GPT~5.5.

As discussed in \cref{sec:ablations}, this higher token expenditure correlates with GPT~5.5's stronger performance on prior-informed benchmarks (\cref{fig:mf:heck} to \cref{fig:mf:dionis}), suggesting a meaningful trade-off between computational budget and optimization quality, though we do not believe this to be a clear causal relationship.
In contrast, Opus~4.8 achieves comparable or superior performance on no-prior benchmarks with fewer tokens, indicating a more computationally efficient tool-use pattern when the optimization relies primarily on the surrogate rather than domain reasoning.

\section{Further discussion}
\label{sec:discussion}

We further discuss observations about agentic BO and about agents equipped with tools such as \texttt{bash}.

\subsection{Discussion on \sara with \texttt{bash}-only}
\label{sec:discussion:bash_only}

To isolate the effect of \lenz's surrogate on the search with \sara, we run a baseline that still uses \lenz for bookkeeping but restricts access to its surrogate and acquisition backend.
Crucially, this baseline receives the same system prompt and strategic guidance as the full \sara (\eg to reason between trials, adapt its search, use the full budget).
The only difference is the lack of access to \lenz's surrogate and acquisition machinery.

\begin{figure}[t]
  \centering
  \begin{subfigure}[b]{\textwidth}
    \centering
    \includegraphics[]{assets/plots/legend_model_no_surrogate.pdf}
  \end{subfigure}\hfill
  \centering
  \begin{subfigure}[b]{0.32\textwidth}
    \centering
    \includegraphics[width=\textwidth]{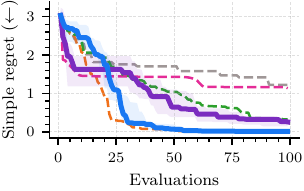}
    \subcaption{Hartmann (6-D)}
    \label{fig:bash_only:hartman}
  \end{subfigure}\hfill
  \begin{subfigure}[b]{0.32\textwidth}
    \centering
    \includegraphics[width=\textwidth]{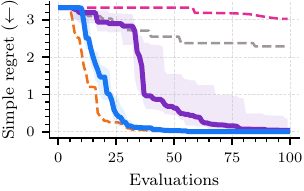}
    \subcaption{const.\ Hartmann (6-D)}
    \label{fig:bash_only:chartmann}
  \end{subfigure}\hfill
  \begin{subfigure}[b]{0.32\textwidth}
    \centering
    \includegraphics[width=\textwidth]{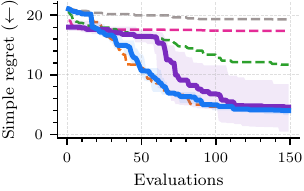}
    \subcaption{Ackley (10-D)}
    \label{fig:bash_only:ackely}
  \end{subfigure}
  \vspace{5pt}
  \begin{subfigure}[b]{0.32\textwidth}
    \centering
    \includegraphics[width=\textwidth]{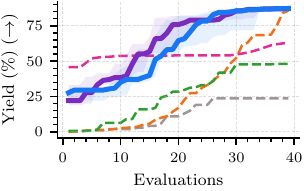}
    \subcaption{Mizoroki--Heck}
    \label{fig:bash_only:heck}
  \end{subfigure}\hfill
  \begin{subfigure}[b]{0.32\textwidth}
    \centering
    \includegraphics[width=\textwidth]{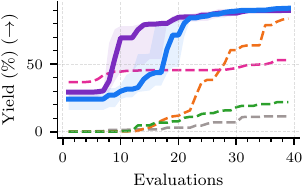}
    \subcaption{Buchwald--Hartwig}
    \label{fig:bash_only:buchwald}
  \end{subfigure}\hfill
  \begin{subfigure}[b]{0.32\textwidth}
    \centering
    \includegraphics[width=\textwidth]{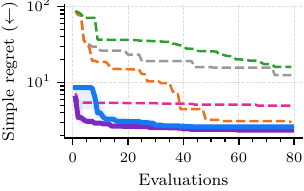}
    \subcaption{LCBench: Dionis}
    \label{fig:bash_only:dionis}
  \end{subfigure}
  \caption{\textit{\texttt{bash} vs.\ \texttt{bash} and \lenz.} Comparison of \sara with access to only \texttt{bash} (no surrogate) against \sara with the full \lenz backend across synthetic and real-world benchmarks.}
  \label{fig:bash_only}
\end{figure}

The results in \cref{fig:bash_only} reveal several notable findings.
First, even without access to the surrogate, \sara with \texttt{bash}-only already achieves stronger performance than LLAMBO and Centaur.
This clearly indicates the importance of having an agent in the loop that can \emph{reason} across different trials, as opposed to simply generating a next configuration based on a summary of past evaluations.
The agent's ability to reflect on the trajectory of observations, form hypotheses about the objective landscape, and adaptively choose its next query provides a qualitative advantage over prompt-based generation approaches.

Second, the value of the surrogate depends on the structure of the objective.
On the coupled synthetic landscapes (Hartmann, constrained Hartmann, and Ackley), the full \lenz backend outperforms \texttt{bash}-only and converges more reliably.
The interactions between dimensions cannot be resolved easily by optimizing each coordinate independently, so the calibrated surrogate, which models the joint response surface, provides a clear edge.
On the reaction-yield and LCBench tasks, by contrast, the two variants reach comparable final performance, and \texttt{bash}-only even converges faster on two of them.
We attribute this to the approximate \emph{separability} of these objectives: the response is dominated by a few largely independent main effects (\eg pH and temperature; \cf~\cref{app:exp-design}), so per-dimension optimization (\cf\ the coordinate-descent strategy below) is near-optimal and the surrogate offers little marginal gain.

\paragraph{Benchmark recognition and evaluation.}
For the synthetic functions we observe distinct ``switching'' points in the optimization trajectories, which approximately correspond to the point at which the agent recognizes and begins to exploit the underlying benchmark, inferring it from the pattern of past evaluations despite the applied shift.
Across seeds, the agent explicitly identified the correct function (by name or by probing its textbook optimum) in \textbf{4/10} runs for Hartmann (6-D), \textbf{3/10} for constrained Hartmann (6-D), and \textbf{10/10} for Ackley (10-D).
Because the optimum was shifted, it could not one-shot the solution; instead it used this structural knowledge to steer the search, \eg the local quadratic fitting described in strategy~3 below.
Once recognized, the agent exploits this knowledge to converge
considerably faster.
However, this observation also exposes a critical challenge for future benchmarking.
The very fact that the agent can recognize a standard test function from a handful of evaluations underlines the fact that popular synthetic benchmarks (Hartmann, Ackley, Branin, and others implemented, \eg, in BoTorch) are clearly part of the pretraining corpus of modern LLMs.
Shifting or rescaling these functions provides only a thin veil; as our results demonstrate, the agent can see through such transformations after sufficiently many observations.
This partly undermines the validity of these benchmarks for evaluating LLM-based optimization: strong performance may reflect \emph{memorization} of known optima rather than genuine optimization capability.
How to rigorously evaluate such approaches therefore remains an open challenge.
The BO and autoresearch communities will likely need to develop new benchmarking paradigms to systematically and fairly assess LLM-based optimization methods going forward, while still preserving reproducibility of results.

Fourth, the \texttt{bash}-only variant exhibits a markedly wider spread of
outcomes across seeds than the surrogate-augmented \sara.
We attribute this to the stochasticity of the agent's reasoning.
Without the stabilizing effect of a fitted surrogate and acquisition function, the trajectory depends more heavily on the agent's initial hypotheses and its exploration--exploitation choices.

We further find that with \texttt{bash} as a tool, the agent can become remarkably creative in attempting to solve the optimization task.
We mainly observe four patterns:
\begin{enumerate}
    \item \textbf{Space-filling initialization:} Lacking the surrogate's Sobol proposals, the agent constructs its own initial design, usually probing corners, the center (also observed in \citet{schwanke2025improving}), and coarse grids or random Latin-hypercube batches implemented in \texttt{bash}, to gauge the objective's scale and locate promising regions before switching to local exploitation.
    We observe this opening consistently on the synthetic benchmarks (\eg \cref{fig:bash_only:hartman}--\cref{fig:bash_only:ackely}), where the exploratory phase precedes the ``switching'' point discussed above.
    \item \textbf{Coordinate descent:} The agent optimizes one variable at a time based on its inferred importance of each dimension.
    This procedure is especially effective for problems whose structure permits such decomposition, such as the chemistry benchmarks presented in this paper (\cf \cref{app:exp-design}), where certain reaction parameters (\eg pH, temperature) dominate the response surface.
    We observe this behavior prominently in \cref{fig:bash_only:heck,fig:bash_only:buchwald}.
    \item \textbf{Local quadratic surrogates:} Rather than moving one coordinate at a time, the agent probes the local curvature of each dimension, fits a quadratic to the resulting values, and combines the per-coordinate minima into a single joint step---a lightweight, hand-built surrogate that it re-fits as the incumbent moves.
    We observe this on the synthetic problems, most prominently in higher dimensions (\eg \cref{fig:bash_only:ackely}), to an extent mimicking approaches such as BayeSQP~\citep{brunzema2025bayesqp}.
    It sits between the model-free coordinate descent above and the full BO loop below and in itself is a variant of surrogate-augmented autoresearch even though the surrogate is self-constructed.
    \item \textbf{Custom-built BO loops:} On the constrained Hartmann problem (\cref{fig:bash_only:chartmann}), we observe that the agent implemented its own GP class and a full BO loop entirely within \texttt{bash}.
    Since we do not provide access to \texttt{numpy} or any scientific computing library, the agent rebuilt all necessary linear algebra operations using only the \texttt{math} package and native Python lists.
    This is both technically impressive and indicative of the agent's capacity for complex, multi-step tool use.
\end{enumerate}

That the agent sometimes rebuilds a GP and acquisition function by hand is itself informative: a calibrated surrogate is not machinery we impose on the agent, but machinery a capable agent reaches for when the problem rewards it.
Providing it natively through \lenz removes a brittle, budget-consuming reinvention step, reduced the variance of the search strategies used by the agent, and lets the agent spend its reasoning on \emph{where} to search rather than on re-deriving \emph{how} to model.

These findings further demonstrate how capable LLMs are in manipulating \texttt{bash} for complex computational tasks.
We therefore see a promising direction in allowing an agentic BO agent to not only passively consume surrogate predictions but also actively modify and extend the optimization infrastructure---for instance, by defining custom priors on the acquisition function or modifying the GP prior (\eg its mean function).

\begin{figure}[t]
  \centering
  \begin{subfigure}[b]{\textwidth}
    \centering
    \includegraphics[]{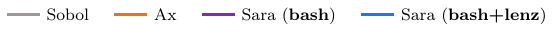}
  \end{subfigure}\hfill
  \centering
  \begin{subfigure}[b]{0.24\textwidth}
    \centering
    \includegraphics[width=\textwidth]{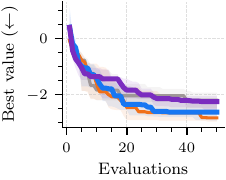}
    \subcaption{GP sample (4-D)}
    \label{fig:gp:gp_sample4}
  \end{subfigure}\hfill
  \begin{subfigure}[b]{0.24\textwidth}
    \centering
    \includegraphics[width=\textwidth]{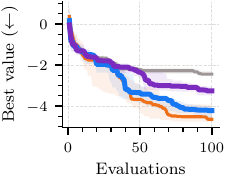}
    \subcaption{GP sample (8-D)}
    \label{fig:gp:gp_sample8}
  \end{subfigure}\hfill
  \begin{subfigure}[b]{0.24\textwidth}
    \centering
    \includegraphics[width=\textwidth]{assets/plots/gp_dims_gp_sample_12d.pdf}
    \subcaption{GP sample (12-D)}
    \label{fig:gp:gp_sample12}
  \end{subfigure}\hfill
  \begin{subfigure}[b]{0.24\textwidth}
    \centering
    \includegraphics[width=\textwidth]{assets/plots/gp_dims_gp_sample_16d.pdf}
    \subcaption{GP sample (16-D)}
    \label{fig:gp:gp_sample16}
  \end{subfigure}
  \caption{\textit{Optimization of GP sample paths.}
  On these benchmarks, the surrogate helps the agent explore an unknown search space systematically.}
  \label{fig:gp}
\end{figure}

\paragraph{Optimization of GP sample paths}

The recognition phenomenon discussed above raises the question of how much of the agent's strong performance on synthetic benchmarks is due to genuine optimization capability versus memorization of known functions.
To evaluate \sara in a setting where function recognition is impossible and the landscape is truly unknown, we optimize randomly generated GP sample paths constructed as described in~\cref{app:gp-samples}.
The resulting functions are by construction absent from any pretraining corpus, exhibit genuine multi-modality, and couple all input dimensions jointly through the kernel's covariance structure, so that coordinate-wise strategies cannot exploit separability.
We vary the dimensionality from 4 to 16 to stress-test scalability.

The results in \cref{fig:gp} reveal a clear and consistent picture: in this pure black-box regime the surrogate-augmented \sara substantially outperforms the \texttt{bash}-only variant across all dimensionalities.
Without recognizable structure to exploit, the agent can no longer rely on ad-hoc strategies such as coordinate descent or benchmark-specific shortcuts; instead, it must systematically explore a high-dimensional, correlated landscape---precisely the setting where a calibrated surrogate and principled acquisition function provide the greatest marginal value.

\subsection{Tool usage}
\label{sec:discussion:tool_usage}

\begin{figure}[t]
  \centering
  \begin{subfigure}[b]{\textwidth}
    \centering
    \includegraphics[width=\textwidth]{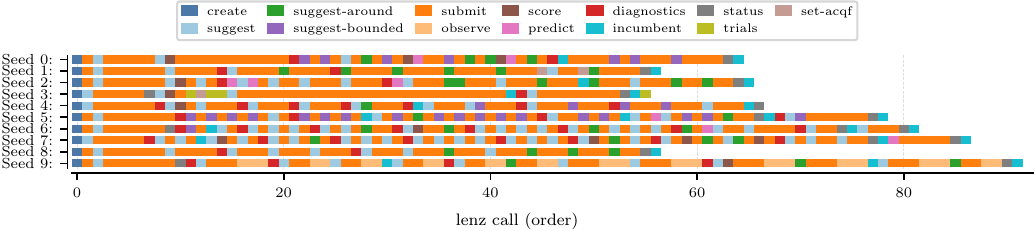}
    \subcaption{Mizoroki--Heck, with prior}
    \label{fig:tool-usage:heck-prior}
  \end{subfigure}

  \begin{subfigure}[b]{\textwidth}
    \centering
    \includegraphics[width=\textwidth]{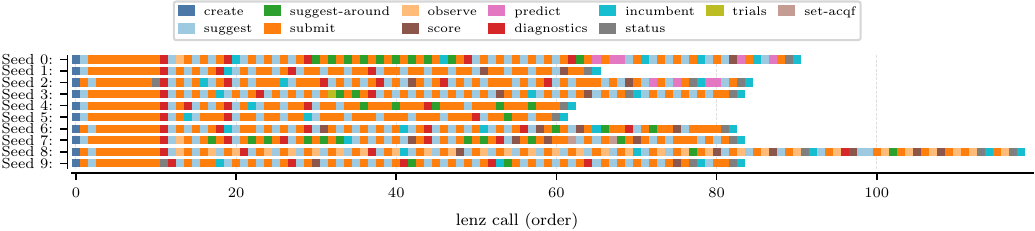}
    \subcaption{Mizoroki--Heck, without prior}
    \label{fig:tool-usage:heck-noprior}
  \end{subfigure}

    \begin{subfigure}[b]{\textwidth}
    \centering
    \includegraphics[width=\textwidth]{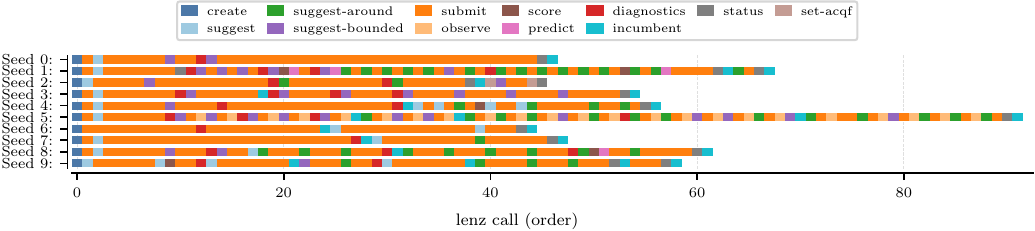}
    \subcaption{Buchwald--Hartwig, with prior}
    \label{fig:tool-usage:buchwald-prior}
  \end{subfigure}

  \begin{subfigure}[b]{\textwidth}
    \centering
    \includegraphics[width=\textwidth]{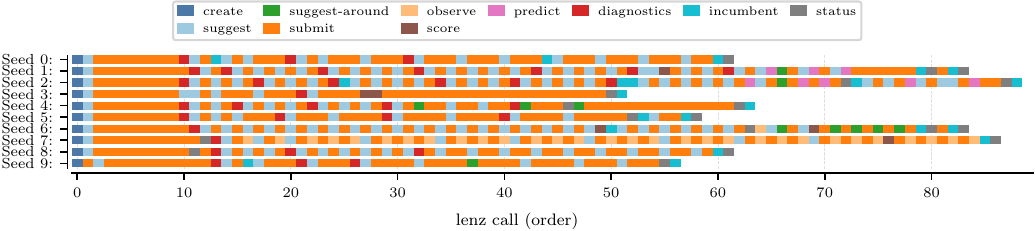}
    \subcaption{Buchwald--Hartwig, without prior}
    \label{fig:tool-usage:buchwald-noprior}
  \end{subfigure}

  \begin{subfigure}[b]{\textwidth}
    \centering
    \includegraphics[width=\textwidth]{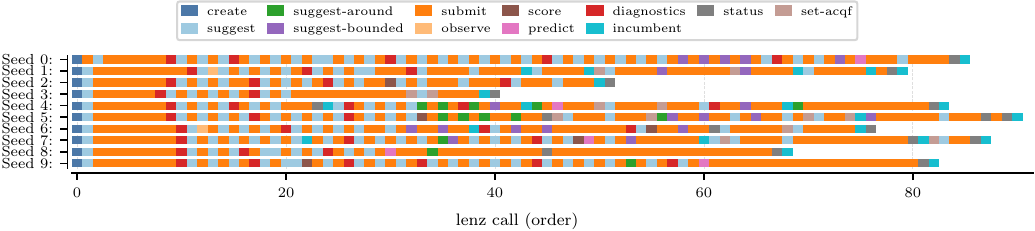}
    \subcaption{Branin}
    \label{fig:tool-usage:branin}
  \end{subfigure}
  \caption{\textit{Tool-use timelines across optimization runs.}
  Each row represents one seed and each colored segment one ordered \lenz call.
  Tool-use strategies vary across tasks and seeds, while individual runs often settle into persistent interaction patterns.}
  \label{fig:tool-usage-timelines}
\end{figure}

Beyond the \texttt{bash}-only analysis, we observe rich and diverse tool-usage patterns across runs with the full \sara system as already demonstrated in \cref{sec:exp:flexibility}.
\Cref{fig:tool-usage-timelines} shows the ordered sequence of \lenz calls for each seed on Mizoroki--Heck and Buchwald--Hartwig, each with and without a prior, and on Branin.
Each row is one optimization run, and each colored segment is one call to a \lenz command, making both the composition and ordering of the agent's deliberation visible.
Across all traces, the agent invokes \texttt{diagnostics}, indicating that it actively inspects surrogate fit and sensitivity information to assess how much to rely on the surrogate rather than treating its proposals as an unquestioned optimizer.

Tool-use patterns differ between tasks, prior conditions, and seeds.
On Heck without a prior, many runs exhibit a regular alternation between a small set of commands, whereas the prior-informed runs show more heterogeneous call sequences.
Branin displays substantial variation across seeds, with some runs maintaining regular patterns and others containing extended blocks of the same command.
Across all five settings, both the total number and ordering of calls vary considerably.
These timelines characterize how \sara uses its tools, but they do not reveal why a command was selected or imply that more tool calls cause better optimization performance.

\section{A meta-MDP completion of agentic BO}
\label{app:meta-mdp}

The metalevel decision process in~\cref{sec:agentic} specifies the information and actions available to an agentic BO policy.
Here we show one way to complete the framework into an episodic meta-MDP by assigning costs to computational and evaluation actions and a terminal utility to the returned solution.

Let $c_{\mathrm{comp}}(a)$ denote the computational cost of a computational action (measured in tokens and backend compute), and let $c_{\mathrm{eval}}(\vx)$ denote the cost of evaluating~$\vx$ (e.g., the compute or laboratory resources consumed by a single experiment).
Let $U(\gD_t,\gK_t)$ denote the terminal utility of the solution returned after~$t$ evaluations under the current requirements.
For example, $U$ may be the incumbent feasible value in a single-objective problem or the attained feasible hypervolume in a multi-objective problem.
For an episode terminating after~$\tau$ evaluations, one possible metalevel objective is
\begin{equation}
    \label{eq:metalevel-objective}
    \max_{\agent}\;
    \E_{\agent}\!\left[
        U(\gD_{\tau},\gK_{\tau})
        - \lambda_{\mathrm{eval}} \sum_{t=0}^{\tau-1} c_{\mathrm{eval}}(\vx_{t+1})
        - \lambda_{\mathrm{comp}} \sum_{t=0}^{\tau} \sum_{j=0}^{k_t-1} c_{\mathrm{comp}}(a_t^{(j)})
    \right],
\end{equation}
subject to any hard evaluation budget, time budget, or resource budget imposed on the campaign.
The weights $\lambda_{\mathrm{eval}}$ and $\lambda_{\mathrm{comp}}$ determine how the policy trades final solution quality against physical experiments and internal deliberation.
Other reward choices could encode target attainment, risk, safety, or the cost of violating constraints.

Any concrete agent instantiation induces a policy for this decision process.
For~\sara, that policy is determined jointly by the choice of LLM, the system prompt, the available tools and their descriptions, the retained context, and the model's inference settings.
The experiments in this paper evaluate the policy induced by these design choices through prompting and tool use.
They do not imply that this policy is optimal for~\cref{eq:metalevel-objective}, or even that the cost weights defining such an optimum are known.

The meta-MDP formulation above could instead provide a training objective for a specialized agentic BO model.
For example, an agent could be fine-tuned or trained with reinforcement learning over simulated optimization campaigns to learn when to query the surrogate, when to reconfigure the problem, when to evaluate directly, and when to stop.
Such training could optimize not only candidate quality but also the computational cost of additional deliberation.
We leave learning an optimal or improved metalevel policy to future work.

\clearpage
\section{\sara system prompt}
\label{app:system-prompt}

For reproducibility and to make \sara's intended behavior fully transparent, we state the complete system prompt below.
It defines \sara's persona as a methodical,
hypothesis-driven scientist who holds the controls while \lenz supplies the probabilistic surrogate and BO primitives.
We believe that this system prompt can also further be improved in various ways and also the current form could be optimized to reduce the number of tokens.
Still, as shown in \cref{sec:experiments}, we achieve strong performance across various tasks and settings as well as model families.
It contains no task-specific knowledge about the experimental benchmarks; problem descriptions are supplied separately according to each experimental condition.
For practitioners, we highly recommend tailoring \sara to their specific use case if possible to always be able to leverage all available prior knowledge as well as to encode certain desired behavior.
A useful practice here is to ask \sara herself how to best frame specific requirements in her \texttt{SYSTEM.md}.

{\footnotesize
\tcbinputlisting{
    breakable,
    colback=metabg,
    colframe=metabg,
    boxrule=0.0pt,
    arc=5pt,
    left=3pt, right=3pt, top=3pt, bottom=3pt,
    listing only,
    listing file=assets/sara_system.md,
    listing options={
        basicstyle=\footnotesize\ttfamily,
        breaklines=true,
        columns=fullflexible,
    }
}
}

\clearpage
\section{\lenz reference sheet}
\label{app:lenz-prompt}

Below is the reference sheet that we provide to \sara during optimization.
We hope that this reference sheet, together with the I/O details in \cref{app:lenz-ref}, will allow others to quickly reimplement the backend interface.

{\footnotesize
\tcbinputlisting{
    breakable,
    colback=dashOrange!10,
    colframe=dashOrange!10,
    boxrule=0.0pt,
    arc=5pt,
    left=3pt, right=3pt, top=3pt, bottom=3pt,
    listing only,
    listing file=assets/lenz_reference.md,
    listing options={
        basicstyle=\footnotesize\ttfamily,
        breaklines=true,
        columns=fullflexible,
    }
}
}

\clearpage
\section{The \lenz command-line interface}
\label{app:lenz-ref}

Every \lenz command prints a single JSON line wrapped in a uniform envelope (\texttt{\{"ok", "command", "result"\}}).
The examples below omit this envelope and show the \texttt{result} payload, grouped according to the four interface roles introduced in~\cref{sec:lenz}.
Values are illustrative.
When outcome constraints are present, \cmd{predict} reports
\texttt{prob\_feasible}, \ie, the posterior probability that a candidate satisfies the constraints, \eg\ $\mathbb{P}(\texttt{flops} \le \text{budget})$ under the constraint's GP posterior.

\begin{minipage}{\textwidth}
\paragraph{\highlight{create}.}
The setup call initializes the search space, objectives, constraints, and acquisition function, and returns the resulting configuration summary.
\par
\centering

\begin{lenzex}
\begin{lstlisting}[style=lenzconsole]
lenz create \
  --space '{"n_layers":{"kind":"range","lower":4,"upper":24,"type":"int"},"d_model":{"kind":"range","lower":256,"upper":1024,"type":"int"},"lr":{"kind":"range","lower":0.00001,"upper":0.01,"log_scale":true}}' \
  --objectives '{"nll":"minimize"}' \
  --constraints '[{"metric":"flops","upper":500}]' \
  --acqf noisy_logei
\end{lstlisting}
\tcblower
\begin{lstlisting}[style=lenzjson]
{
  "space": {
    "n_layers": {"kind": "range", "lower": 4, "upper": 24, "type": "int"},
    "d_model": {"kind": "range", "lower": 256, "upper": 1024, "type": "int"},
    "lr": {"kind": "range", "lower": 0.00001, "upper": 0.01, "log_scale": true}
  },
  "objectives": [{"metric": "nll", "minimize": true}],
  "constraints": [{"metric": "flops", "lower": null, "upper": 500}],
  "acqf": "noisy_logei",
  "is_moo": false
}
\end{lstlisting}
\end{lenzex}

\captionof{figure}{Creating a \lenz study. \cmd{create} materializes the initial optimization configuration and returns a summary that the agent can check against the intended problem.}
\label{fig:lenz-create}
\end{minipage}

\begin{minipage}{\textwidth}
\paragraph{\highlight{propose} and \highlight{evaluate}  {\normalfont(\cmd{suggest}, \cmd{submit}, \cmd{observe})}.}
\suggest\ returns one or more candidates without modifying the trial log.
The agent may adopt, refine, or override a proposal; once it commits a candidate, \cmd{submit} marks it in flight and \cmd{observe} attaches the resulting experimental metrics.
\par
\centering

\begin{lenzex}
\begin{lstlisting}[style=lenzconsole]
lenz suggest
\end{lstlisting}
\tcblower
\begin{lstlisting}[style=lenzjson]
[
  {
    "config": {"n_layers": 12, "d_model": 768, "lr": 0.0003},
    "acquisition_values": {"logei": -1.92},
    "trial_id": null,
    "acqf": "logei"
  }
]
\end{lstlisting}
\end{lenzex}

\captionof{figure}{Proposing with \lenz. \suggest\ returns a candidate and its acquisition value but records nothing, as indicated by the \texttt{null} \texttt{trial\_id}. A real evaluation enters the trial log only through \cmd{submit} and \cmd{observe}.}
\label{fig:lenz-propose}
\end{minipage}

\begin{minipage}{\textwidth}
\paragraph{\highlight{reconfigure}  {\normalfont(\cmd{set-bounds}, \cmd{set-acqf}, \cmd{set-objectives}, \cmd{set-constraints})}.}
These commands persistently edit the live optimization specification and return its updated summary without discarding observations.
\par
\centering

\begin{lenzex}
\begin{lstlisting}[style=lenzconsole]
lenz set-acqf --acqf ucb
\end{lstlisting}
\tcblower
\begin{lstlisting}[style=lenzjson]
{"acqf": "ucb", "is_moo": false, "observed": 12}
\end{lstlisting}
\end{lenzex}

\vspace{2pt}

\begin{lenzex}
\begin{lstlisting}[style=lenzconsole]
lenz set-bounds --bounds '{"lr": [0.0001, 0.001]}'
\end{lstlisting}
\tcblower
\begin{lstlisting}[style=lenzjson]
{"bounds": {"lr": [0.0001, 0.001]}, "is_moo": false, "observed": 12}
\end{lstlisting}
\end{lenzex}

\vspace{2pt}

\begin{lenzex}
\begin{lstlisting}[style=lenzconsole]
lenz set-objectives --objectives '{"nll":"minimize","flops":"minimize"}'
\end{lstlisting}
\tcblower
\begin{lstlisting}[style=lenzjson]
{
  "is_moo": true,
  "objectives": [
    {"metric": "nll", "minimize": true},
    {"metric": "flops", "minimize": true}
  ]
}
\end{lstlisting}
\end{lenzex}

\vspace{2pt}

\begin{lenzex}
\begin{lstlisting}[style=lenzconsole]
lenz set-constraints --constraints '[{"metric":"flops","upper":500}]'
\end{lstlisting}
\tcblower
\begin{lstlisting}[style=lenzjson]
{
  "constraints": [{"metric": "flops", "lower": null, "upper": 500}],
  "is_moo": false
}
\end{lstlisting}
\end{lenzex}

\captionof{figure}{Reconfiguring a \lenz study. Each command edits the live problem and returns the updated configuration summary, allowing \sara to change the search strategy or problem specification mid-run.}
\label{fig:lenz-reconfigure}
\end{minipage}

\begin{minipage}{\textwidth}
\paragraph{\highlight{probe} {\normalfont (\cmd{predict}, \cmd{score}, \cmd{diagnostics}, \cmd{incumbent}, \cmd{pareto}, \cmd{trials}, \cmd{status})}.}
Probe calls are read-only. 
They expose posterior predictions, acquisition utilities, diagnostics, and campaign state without changing the configuration.
\par
\centering

\begin{lenzex}
\begin{lstlisting}[style=lenzconsole]
lenz predict --configs '[{...}]'
\end{lstlisting}
\tcblower
\begin{lstlisting}[style=lenzjson]
[
  {
    "mean": {"nll": 2.41, "flops": 462.0},
    "variance": {"nll": 0.006, "flops": 900.0},
    "prob_feasible": 0.90,
    "acquisition_values": {}
  }
]
\end{lstlisting}
\end{lenzex}

\vspace{2pt}

\begin{lenzex}
\begin{lstlisting}[style=lenzconsole]
lenz score --acqf logei,ucb --configs '[{...}, {...}]'
\end{lstlisting}
\tcblower
\begin{lstlisting}[style=lenzjson]
[
  {"logei": -1.92, "ucb": 0.44},
  {"logei": -3.05, "ucb": 0.10}
]
\end{lstlisting}
\end{lenzex}

\vspace{2pt}

\begin{lenzex}
\begin{lstlisting}[style=lenzconsole]
lenz diagnostics
\end{lstlisting}
\tcblower
\begin{lstlisting}[style=lenzjson]
{
  "objective": "nll",
  "n_observed": 12,
  "cv_r2": 0.88,
  "noise": 0.0035,
  "lengthscales": {"n_layers": 0.42, "d_model": 0.30, "lr": 0.09},
  "sensitivity": {"n_layers": 0.22, "d_model": 0.34, "lr": 0.55}
}
\end{lstlisting}
\end{lenzex}

\captionof{figure}{Probing a \lenz study. \cmd{predict} returns posterior moments and feasibility probabilities, \cmd{score} compares agent-supplied candidates by acquisition utility, and \cmd{diagnostics} exposes fit quality, hyperparameters, and sensitivities.}
\label{fig:lenz-probe}
\end{minipage}

\end{document}